\documentclass[letterpaper]{article} % DO NOT CHANGE THIS
\usepackage{aaai2027}
\nocopyright
\usepackage[hyphens]{url}  % DO NOT CHANGE THIS
\usepackage{graphicx} % DO NOT CHANGE THIS
\usepackage{natbib}  % DO NOT CHANGE THIS AND DO NOT ADD ANY OPTIONS TO IT
\usepackage{caption} % DO NOT CHANGE THIS AND DO NOT ADD ANY OPTIONS TO IT
\usepackage{booktabs}
\usepackage{multirow}
\usepackage{amsmath,amssymb}
\usepackage{array}

\newcolumntype{C}[1]{>{\centering\arraybackslash}m{#1}}
\newcolumntype{L}[1]{>{\raggedright\arraybackslash}m{#1}}
\newcommand{\appTableStyle}{%
    \small
}

\title{CrossScope: A Role-Asymmetric World Model for Joint Dual-Scope Surgical Video Prediction}
\author{
    Wanhao Liu\textsuperscript{\rm 2,\rm 3}\equalcontrib,
    Jinsong Lin\textsuperscript{\rm 1}\equalcontrib,
    Rulin Zhou\textsuperscript{\rm 1}\equalcontrib,
    CHI KIT NG\textsuperscript{\rm 1}\equalcontrib,
    Wenbin Pan\textsuperscript{\rm 2},
    Zhiqing Tang\textsuperscript{\rm 2},
    Dongyue Li\textsuperscript{\rm 2},
    Liwei Luo\textsuperscript{\rm 2},
    Wu Yanshen\textsuperscript{\rm 2},
    Panshuo Li\textsuperscript{\rm 2},
    Zhiyong Xiong\textsuperscript{\rm 4},
    Huxin Gao\textsuperscript{\rm 1},
    Tamas Haidegger\textsuperscript{\rm 5},
    Hongliang Ren\textsuperscript{\rm 1,\rm 3}\corresponding
}
\affiliations{
    \textsuperscript{\rm 1}CUHK,
    \textsuperscript{\rm 2}GDUT,
    \textsuperscript{\rm 3}SLAI,
    \textsuperscript{\rm 4}SYSU,
    \textsuperscript{\rm 5}IROB
}

\begin{document}

\maketitle

\begin{abstract}
Visual world models typically learn future dynamics from a single observation stream, limiting their ability to model cooperative systems with multiple independently moving observers. We investigate this challenge in Mother--Child endoscopic retrograde cholangiopancreatography (ERCP), where two flexible scopes provide complementary yet role-dependent views without a calibrated stereo relationship. Unlike conventional multi-view fusion that assumes symmetric information exchange, we formulate \textbf{role-asymmetric dual-scope future prediction}, where cross-view evidence is selectively transferred according to the prediction target and its underlying spatial requirements. We propose \textbf{CrossScope}, a dual-stream surgical world model that preserves view-specific experts while enabling target-specific evidence routing through geometry-guided residual interactions. CrossScope learns two complementary communication directions: geometric motion cues from the Mother view guide Child-view future dynamics, while pose-aligned Child appearance supports Mother-view prediction only when valid spatial correspondence is established. This design allows each scope to contribute task-relevant evidence without compromising its view-specific representation. To evaluate this problem, we establish a paired dual-scope benchmark comprising synchronized phantom and real-world ERCP episodes, with evaluations assessing visual fidelity, structural preservation, target localization, and motion consistency. Experiments demonstrate that CrossScope consistently outperforms strong surgical video generation baselines, validating the importance of role-aware evidence routing for multi-observer visual world modeling.
\end{abstract}

% Add final public links here if applicable.

\section{Introduction}

% Visual world models forecast future observations from visual states \cite{ha2018worldmodels,bruce2024genie}. SurgVista targets long-horizon surgical dynamics and Cosmos-H-Surgical connects video prediction to policy learning \cite{pan2026surgvista,he2025cosmossurgical}, but neither models role-dependent exchange across independently actuated scopes. Here, world model denotes observation-conditioned visual dynamics rather than action-conditioned counterfactual rollouts.

Visual world models aim to learn observation dynamics and predict future states of an environment from partial observations \cite{ha2018worldmodels,bruce2024genie}. Recent advances in surgical world models have demonstrated promising progress in forecasting surgical video dynamics, including long-horizon predictions that capture instrument--tissue interactions \cite{pan2026surgvista}. However, existing approaches primarily focus on a single visual stream, implicitly assuming that future dynamics can be inferred from one observer. This assumption becomes insufficient for cooperative procedures involving multiple independently controlled views, where different observers may provide complementary but non-equivalent evidence.

\begin{figure}[t]
    \centering
    \includegraphics[width=\linewidth]{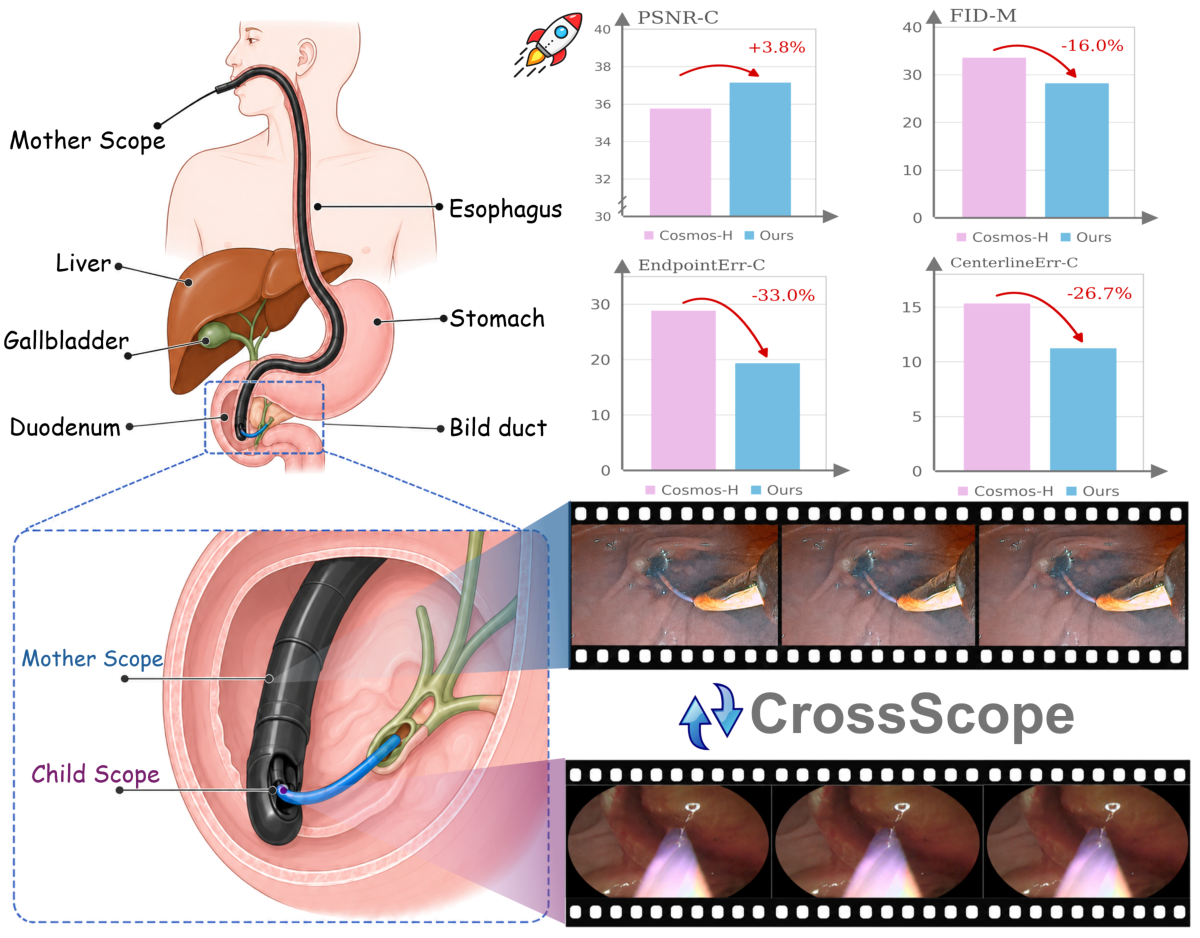}
    \caption{CrossScope for role-asymmetric dual-scope future prediction.
    Mother and Child scopes provide complementary wide- and near-field observations; directional M2C and C2M routes jointly forecast both streams. The plots show representative phantom-benchmark gains over Cosmos-H-Surgical.}
    \label{fig:teaser}
\end{figure}

Mother--Child endoscopic retrograde cholangiopancreatography (ERCP) presents a distinct dual-scope prediction problem. The wide-view Mother duodenoscope provides global spatial context and visualizes the Child-scope tip, whereas the near-field Child cholangioscope advances through the Mother into narrow bile ducts to reveal local anatomy \cite{parsi2011peroral}. The two flexible scopes are independently controlled rather than fixed as a calibrated stereo pair. Their relative pose, image scale, spatial overlap, and visible content therefore vary throughout a procedure. The challenge is therefore not to treat the two views as interchangeable inputs, but to determine which cross-scope evidence can condition each prediction target. We formulate this setting as \emph{role-asymmetric dual-scope future prediction}: jointly forecasting both streams from current anchors and strictly causal Child history.

Yet existing surgical video methods do not model role-dependent exchange across independently moving scopes. Conditional generators enrich one target with visual, object-level, structured-scene, or action cues \cite{li2024endora,iliash2024interactive,chen2025surgsora,biagini2025hierasurg,yeganeh2024visage,sivakumar2025sg2vid,koju2025surgical}, while SAW and Routed Visual Control add trajectories or image-aligned kinematics \cite{rapuri2026saw,li2026routed}. These controls improve target-specific prediction but do not specify what evidence should transfer from another independently moving view. Multi-view reconstruction assumes shared geometry or jointly optimized poses \cite{ozsoy2022fourDOR,li2024endosparse,huang2024endo4dgs,stilz2026flex}. Temporal memory retains evidence within one stream or shared representation, while surgical foundation models focus on representation learning \cite{cheng2022xmem,xiao2025worldmem,lu2026surgrec,wu2026surgmotion,yang2026surgvista}. These approaches do not match Mother--Child ERCP: Mother-plane geometry can guide Child motion, whereas Child appearance can inform Mother prediction only after causal observation and spatial alignment.

% To address this gap, we propose CrossScope, which replaces symmetric exchange with target-dependent evidence allocation (Figure~\ref{fig:teaser}). Separate Mother and Child experts expose read-only pre-injection states and exchange only zero-initialized, target-specific residuals. Pose Readout yields a Mother-plane Child-tip trajectory for Mother-to-Child (M2C) motion conditioning; Child-to-Mother (C2M) pose-aligns the observed Child anchor and causal historical observations before writing geometry-supported Mother residuals. To evaluate this formulation, we establish, to our knowledge, the first paired flexible dual-scope benchmark for joint future prediction, comprising phantom and real-world ERCP episodes. Motivated by evaluation work that distinguishes visual realism from higher-level surgical plausibility \cite{chen2025surgveo}, we assess frame fidelity together with role-specific structure, localization, and motion. Complete-system controls, same-backbone route ablations, and a two-model comparison on seven held-out real-world episodes test the proposed allocation within the evaluated cohort.

To address this challenge, we introduce \textbf{CrossScope}, a role-asymmetric world model for joint dual-scope surgical video prediction. Instead of treating multiple views as symmetric observations, CrossScope models cross-scope interaction through target-dependent evidence allocation, allowing each observer to contribute information according to its predictive role. We further establish a paired dual-scope benchmark with synchronized phantom and real-world ERCP episodes to evaluate future prediction beyond visual fidelity, including structural consistency, target localization, and motion accuracy. Comprehensive experiments demonstrate that CrossScope consistently outperforms strong video generation baselines, highlighting the importance of role-aware evidence routing for multi-observer visual world modeling.

Our contributions are fourfold:
\begin{enumerate}
    \item \textbf{Role-asymmetric dual-scope future prediction.} We formulate joint future prediction for Mother--Child ERCP, where current observations and causal history provide directional, non-interchangeable evidence across views.
    \item \textbf{A role-preserving read/write contract.} Separate experts expose pre-injection states through a read-only interface, while cross-scope communication is restricted to zero-initialized, target-specific residual writes.
    \item \textbf{Target-specific geometric conditioning.} M2C derives Child motion conditions from a predicted Mother-plane trajectory; C2M separately translates pose-aligned Current and History, then blends their residuals only where geometry licenses both sources.
    \item \textbf{A paired dual-scope benchmark and protocol.} We provide phantom and real-world episodes with role-specific evaluation of frame fidelity, Mother structure, Child localization, and Child trajectories.
\end{enumerate}

\begin{figure}[ht]
    \centering
    \includegraphics[width=\linewidth]{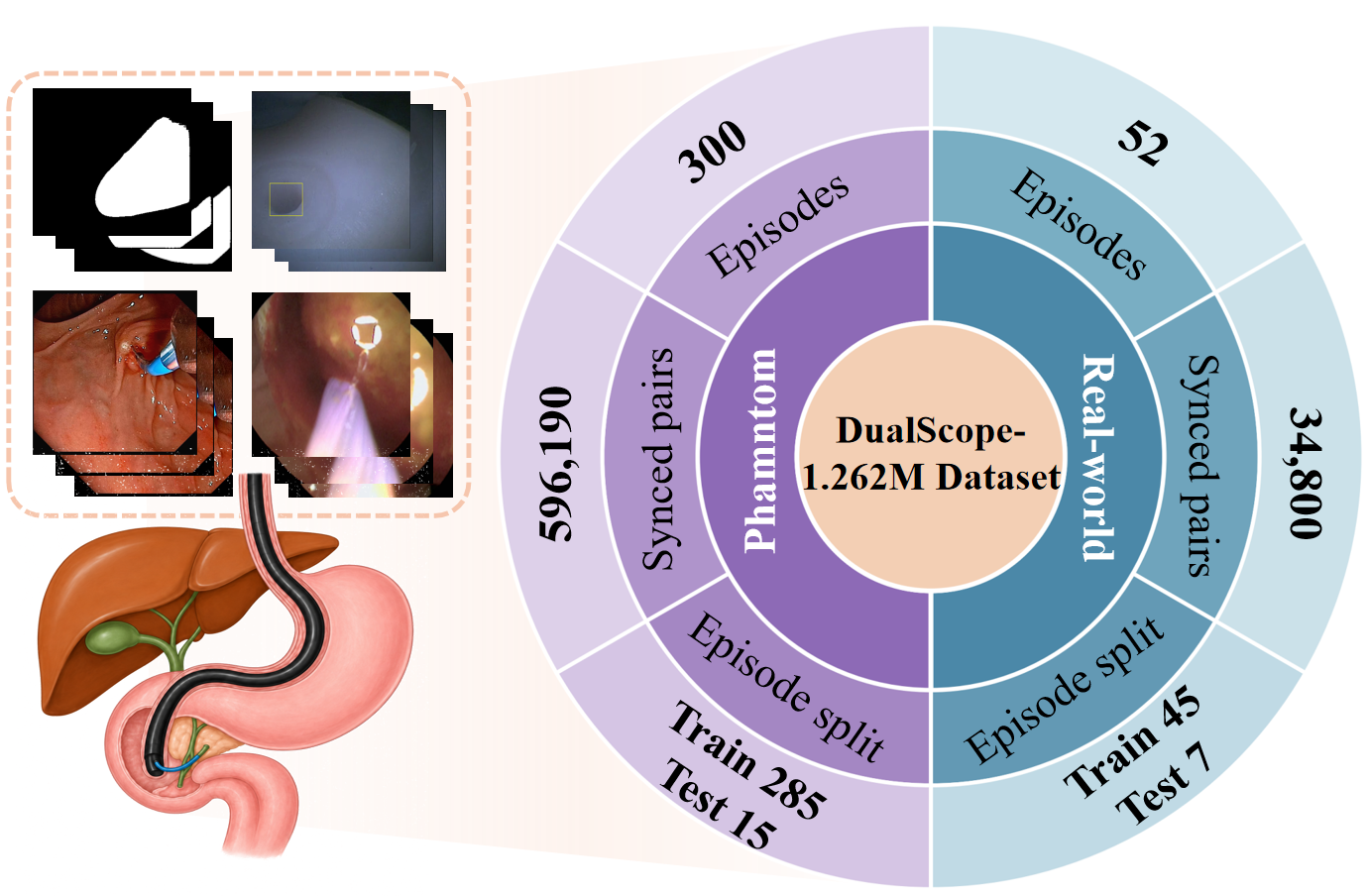}
    \caption{Dual-scope benchmark composition.
    The benchmark comprises paired phantom and real-world Mother--Child ERCP episodes with synchronized wide-view Mother and near-field Child streams; $1.262$M denotes both views counted as individual frames.}
    \label{fig:dataset}
\end{figure}

\begin{figure*}[!t]
    \centering
    \includegraphics[width=\linewidth]{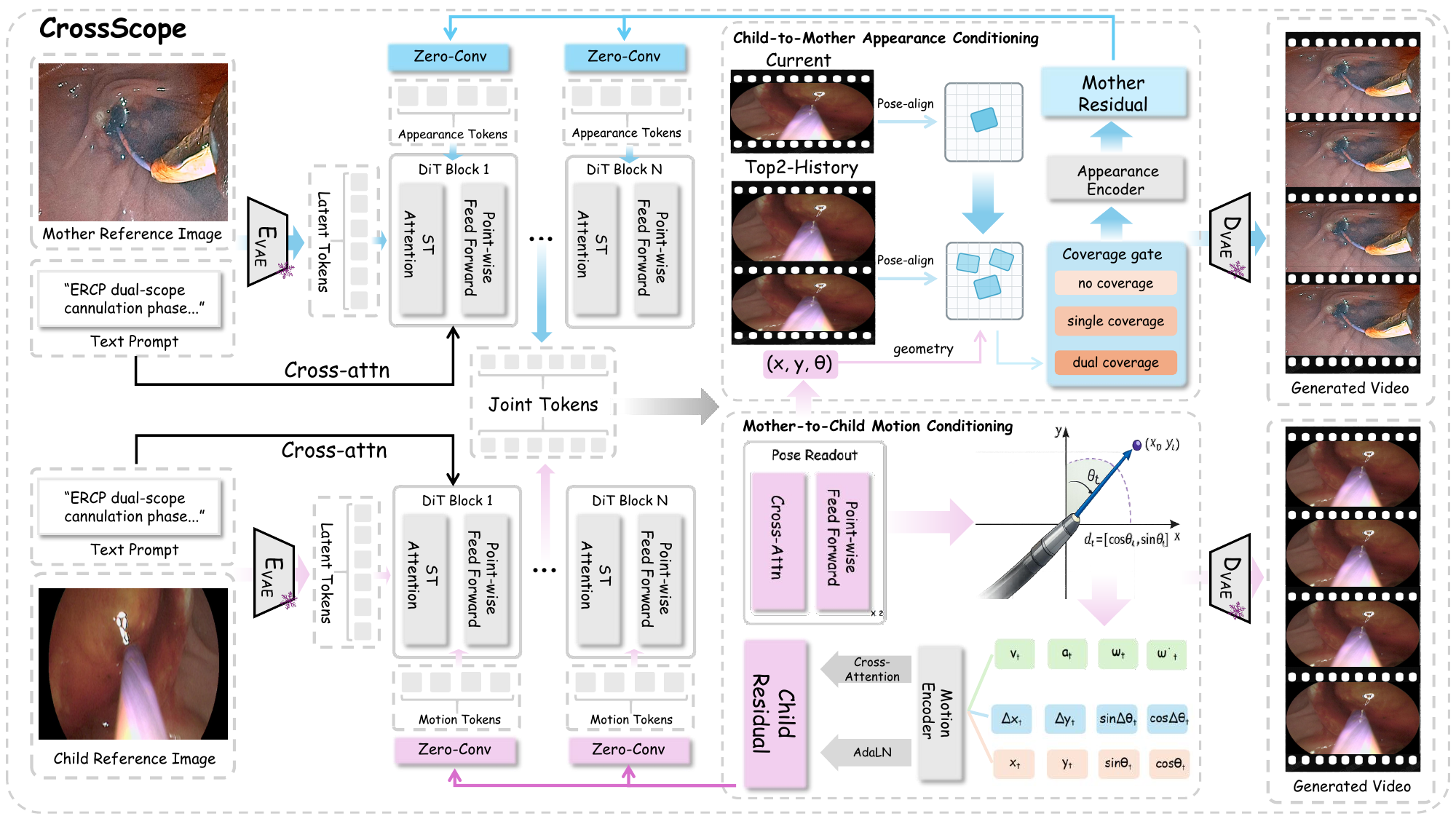}
    \caption{CrossScope architecture.
    Separate DiT experts expose read-only pre-injection states. Frame $0$ is the observed anchor, and future latent groups cover frames $1$--$4$ and $5$--$8$. Pose Readout maps Mother states to a Mother-plane trajectory, which M2C converts into Child motion residuals. C2M pose-aligns Current and up to two causal History observations, then writes or blends Mother residuals only under geometric support.}
    \label{fig:architecture}
\end{figure*}

\section{Dual-Scope Benchmark}
\label{sec:benchmark}

We introduce a \textbf{paired dual-scope benchmark} for evaluating multi-observer visual world models in Mother--Child ERCP (Figure~\ref{fig:dataset}). Unlike conventional single-view surgical video datasets, each episode contains synchronized wide-view Mother and near-field Child streams with independently controlled scope motions.

This setting enables systematic evaluation of future prediction under complementary yet asymmetric observations, where non-shared motion, view-dependent visibility, and role-specific prediction errors can be measured explicitly. The benchmark consists of both phantom and real-world ERCP data. The phantom subset contains 300 paired episodes with 596,190 synchronized time points, including 285 training episodes and 15 test episodes. The real-world subset contains 52 paired episodes with 34,800 synchronized time points, split into 45 training and seven held-out test episodes. All splits are episode-disjoint, and the real-world evaluation further ensures patient- and case-level separation between training and testing.

Both domains provide complementary view-specific annotations, including Mother background/instrument segmentation masks and Child papilla bounding boxes. These annotations enable evaluation across multiple dimensions, including structural preservation in the Mother view, target localization in the Child view, and future motion consistency across both streams. The paired synchronization and annotation protocol were verified by clinicians and are described in Appendix~\ref{app:benchmark}. The retrospective real-world study was conducted under ethics approval with a waiver of informed consent, and all videos were de-identified. The phantom videos, annotations, data splits, and evaluation code will be released, while real-world data will remain available only through controlled access following applicable ethics and data-use requirements.

\section{Method}

CrossScope treats dual-scope interaction as target-dependent evidence allocation rather than symmetric fusion (Figure~\ref{fig:architecture}). Separate Mother and Child DiT experts preserve view-specific statistics, and their pre-injection states form a read-only interface. Pose Readout estimates a Mother-plane Child-tip trajectory from the Mother partition and anchor; M2C and C2M convert relevant evidence into target-specific residuals without altering this snapshot. The Mother coordinate frame locates Child motion but does not make Child appearance globally transferable. Thus, M2C conditions Child motion from the trajectory, whereas C2M writes Mother appearance only where geometry licenses observed Child evidence. We first define the causal read/write contract, then specify M2C and C2M, before describing joint learning and the causal inference schedule.

\subsection{Causal Formulation and Role-Asymmetric Dual-Stream Backbone}

Although synchronized, the independently actuated streams do not form a calibrated stereo pair and cannot be represented by a shared latent with interchangeable evidence. We therefore preserve view-specific experts and permit only explicit, target-directed residual communication. Given synchronized wide-view Mother and near-field Child videos, we predict eight frames after their observed anchors. Let $I_d^{0:8}$ be a nine-frame clip for $d\in\{M,C\}$, $c$ a fixed prompt, and $\mathcal H_C=\{(I_{C,j},\hat p_j,a_j)\}$ a same-episode causal cache of Child observations whose times precede the anchor by at least 33 synchronized time points, with cached predicted Mother-plane poses and ages. We model
\begin{equation}
p_\theta\!\left(I_M^{1:8},I_C^{1:8}\mid
I_M^0,I_C^0,\mathcal H_C,c\right).
\label{eq:causal_prediction}
\end{equation}

The VAE encodes each stream into $Z_d=(z_d^0,z_d^1,z_d^2)$: $z_d^0$ represents frame $0$, $z_d^1$ frames $1$--$4$, and $z_d^2$ frames $5$--$8$. Separate DiT experts process the two streams while conditioning on $c$. At routing depth $r$, their pre-injection states form the read-only Joint Tokens $J^{(r)}=[X_M^{(r)};X_C^{(r)}]$.
For a within-stream update $\bar X_d^\ell=F_d^\ell(X_d^\ell;c)$, cross-scope communication is restricted to
\begin{equation}
\begin{aligned}
X_M^{\ell+1} &= \bar X_M^\ell+R_{C\rightarrow M}^\ell,\\
X_C^{\ell+1} &= \bar X_C^\ell+R_{M\rightarrow C}^\ell.
\end{aligned}
\label{eq:directed_updates}
\end{equation}
Zero-initialized output projections recover the uncoupled experts at initialization. All routing reads come from the pre-injection snapshot, so neither residual depends on the other path's same-layer write.

\subsection{Mother-to-Child Motion Conditioning}

The near-field Child stream lacks a stable reference for global displacement. The lower M2C path therefore resolves Child-motion ambiguity through a Mother-plane Child-tip trajectory. Pose Readout forms queries from the Mother anchor $z_M^0$ and flow-matching timestep embedding $e_\sigma$, then attends to the Mother partition of the pre-injection states:
\begin{equation}
\begin{aligned}
\hat P &= \mathcal R\!\left(X_M^{(r)},z_M^0,e_\sigma\right)
=(\hat p_0,\ldots,\hat p_8),\\
\hat p_t &= \left(\hat x_t,\hat y_t,
\widehat{\cos\theta_t},\widehat{\sin\theta_t},\hat\nu_t\right).
\end{aligned}
\label{eq:pose_readout}
\end{equation}
The resulting $\hat P$ is expressed in the Mother plane, while M2C writes only to the Child expert. Positions lie in $[-0.5,0.5]^2$, $\hat\nu_t$ is the predicted validity score used to mask invalid states, and stop-gradient prevents generation losses from redefining this interface. Appendix~\ref{app:pose} provides quantitative diagnostics and visual overlays.

The Mother partition supplies wide-view localization and the pinned anchor. This shared coordinate convention gives both routes a compact geometric bottleneck without transferring appearance.

Because a single endpoint displacement cannot represent both local turning and accumulated advance, M2C encodes $\operatorname{sg}(\hat P)$ at fine and coarse temporal scales, where $\operatorname{sg}$ denotes stop-gradient and hats are omitted below:
\begin{equation}
\begin{aligned}
\phi_t={}&[x_t,y_t,\cos\theta_t,\sin\theta_t,
\\
&\Delta x_t^{\mathrm{ego}},\Delta y_t^{\mathrm{ego}},
\sin\Delta\theta_t,\cos\Delta\theta_t,\\
&v_t,\omega_t,a_t,\alpha_t]\in\mathbb R^{12},\\
m_g^{\mathrm{coarse}}={}&[\Delta x_g^{\mathrm{ego}},
\Delta y_g^{\mathrm{ego}},\\
&\sin\Delta\theta_g,\cos\Delta\theta_g],
\quad g\in\{1,2\}.
\end{aligned}
\label{eq:motion_descriptors}
\end{equation}
Here $v_t,\omega_t,a_t,\alpha_t$ denote linear speed, angular velocity, linear acceleration, and angular acceleration. The eight descriptors, indexed by $t\in\{0,\ldots,7\}$, encode transitions $(0\!\rightarrow\!1),\ldots,(7\!\rightarrow\!8)$: $z_C^1$ cross-attends to $\phi_{0:3}$ and $z_C^2$ to $\phi_{4:7}$. AdaLN uses the corresponding coarse endpoint conditions $(0,4)$ and $(4,8)$. The fine cross-attention and coarse AdaLN branches produce $R_{\mathrm{attn}}^\ell$ and $R_{\mathrm{AdaLN}}^\ell$, whose sum forms the Child residual
\begin{equation}
R_{M\rightarrow C}^\ell
=\alpha_{\mathrm w}\mathcal Z_C^\ell\!\left(
R_{\mathrm{attn}}^\ell+R_{\mathrm{AdaLN}}^\ell\right).
\label{eq:m2c_residual}
\end{equation}
The two conditions resolve complementary temporal scales: cross-attended $\phi_t$ preserves adjacent displacement and heading changes, while coarse AdaLN summarizes net motion over each four-transition group. Their combination retains transition-level variation while constraining accumulated displacement over the predicted horizon.

Zero-initialized $\mathcal Z_C^\ell$ is enabled through the warmup $\alpha_{\mathrm w}$. To supervise the motion written to the Child expert rather than only the intermediate pose, an auxiliary head decodes a Child box trajectory from layer-averaged residual fields and applies trajectory, endpoint, ROI, and validity losses.

\begin{table*}[!t]
    \centering
    \begingroup
    \small
    \begin{tabular*}{\linewidth}{@{\extracolsep{\fill}}lcccccccc@{}}
        \toprule
        \multicolumn{1}{c}{\multirow{2}{*}{\textbf{Model}}} &
        \multicolumn{4}{c}{\rule{0pt}{2.5ex}\textbf{Child View}} &
        \multicolumn{4}{c}{\rule{0pt}{2.5ex}\textbf{Mother View}} \\
        \cmidrule(lr){2-5}\cmidrule(lr){6-9}
        & \textbf{PSNR}$\uparrow$ & \textbf{SSIM}$\uparrow$ & \textbf{FID}$\downarrow$ & \textbf{LPIPS}$\downarrow$ &
        \textbf{PSNR}$\uparrow$ & \textbf{SSIM}$\uparrow$ & \textbf{FID}$\downarrow$ & \textbf{LPIPS}$\downarrow$ \\
        \midrule
        \multicolumn{9}{@{}c@{}}{\rule[-0.7ex]{0pt}{3.0ex}\textit{(a) Frame Fidelity}} \\
        \addlinespace[1pt]
        HunyuanVideo-I2V & 34.332 & 0.942 & 18.213 & 0.0848 & 35.112 & 0.953 & 34.324 & 0.0352 \\
        Cosmos-H-Surgical & \underline{35.123} & \underline{0.943} & \underline{17.431} & \underline{0.0821} & 35.774 & \underline{0.969} & 33.617 & \underline{0.0347} \\
        Wan2.2 & 35.003 & 0.940 & 17.986 & 0.0854 & 35.001 & 0.956 & 34.618 & 0.0355 \\
        \addlinespace[2pt]
        Early-Fusion Wan2.2 & 30.790 & 0.915 & 20.190 & 0.0877 & 35.432 & 0.966 & 33.378 & 0.0354 \\
        Symmetric Cross-Attention MoT & 33.326 & 0.931 & 18.369 & 0.0852 & \underline{36.298} & 0.966 & \underline{29.019} & 0.0351 \\
        \textbf{CrossScope} & \textbf{35.906} & \textbf{0.951} & \textbf{17.402} & \textbf{0.0819} & \textbf{37.147} & \textbf{0.973} & \textbf{28.232} & \textbf{0.0334} \\
        \bottomrule
    \end{tabular*}

    \begin{tabular*}{\linewidth}{@{\extracolsep{\fill}}lccccc@{}}
        \toprule
        \multicolumn{1}{c}{\multirow{2}{*}{\textbf{Model}}} &
        \multicolumn{1}{c}{\rule{0pt}{2.5ex}\textbf{Mother Structure}} &
        \multicolumn{2}{c}{\rule{0pt}{2.5ex}\textbf{Child Localization}} &
        \multicolumn{2}{c}{\rule{0pt}{2.5ex}\textbf{Child Trajectory}} \\
        \cmidrule(lr){2-2}\cmidrule(lr){3-4}\cmidrule(lr){5-6}
        & \textbf{Mask IoU}$\uparrow$ & \textbf{Box IoU}$\uparrow$ & \textbf{Det. Recall}$\uparrow$ & \textbf{Endpoint Err.}$\downarrow$ & \textbf{Centerline Err.}$\downarrow$ \\
        \midrule
        \multicolumn{6}{@{}c@{}}{\rule[-0.7ex]{0pt}{3.0ex}\textit{(b) Task Fidelity and Motion}} \\
        \addlinespace[1pt]
        HunyuanVideo-I2V & \underline{0.946} & 0.789 & 0.909 & 36.068 & 19.126 \\
        Cosmos-H-Surgical & 0.937 & \underline{0.802} & 0.913 & \underline{28.880} & 15.339 \\
        Wan2.2 & 0.944 & 0.792 & 0.911 & 35.093 & 18.544 \\
        \addlinespace[2pt]
        Early-Fusion Wan2.2 & 0.933 & 0.724 & 0.907 & 42.590 & 21.733 \\
        Symmetric Cross-Attention MoT & 0.941 & 0.798 & \underline{0.917} & 31.767 & \underline{12.414} \\
        \textbf{CrossScope} & \textbf{0.948} & \textbf{0.839} & \textbf{0.927} & \textbf{19.343} & \textbf{11.238} \\
        \bottomrule
    \end{tabular*}
    \endgroup
    \caption{End-to-end comparison on the phantom benchmark. All systems use the common training split and evaluation windows. Panel (a) reports view-specific frame fidelity and panel (b) task fidelity and motion; trajectory errors are in pixels. Bold and underlined values mark the best and second-best results.}
    \label{tab:main_comparison}
\end{table*}

\subsection{Child-to-Mother Appearance Conditioning}

The Child view may reveal local anatomy that is not visible in the wider Mother view, but the observed Child appearance is neither globally registered to the Mother view nor valid as future evidence. C2M therefore establishes Mother-plane support from observed sources before writing any appearance residual. It projects the inference-available anchor $z_C^0$ once with $\hat p_0$ and broadcasts the static canvas to future Mother groups. Historical slices use their cached poses, preventing pose--appearance mismatch. The future trajectory guides ROI selection but contributes no appearance; only observed Current and History latents do. For local Child coordinate $u$ and Mother-plane pose $p=(x,y,\theta)$,
\begin{equation}
\pi(p,u)=
\begin{bmatrix}x\\y\end{bmatrix}
+\operatorname{Rot}(\theta)\big(Au+\delta(u)\big),
\label{eq:pose_align}
\end{equation}
where the global footprint $A$ and bounded correction $\delta(u)$ are learned through Mother flow matching. Only valid in-frame projections write the static canvas, defining spatial support without calibrated stereo geometry or pixel copying.

The current Child anchor alone may not cover all Mother-view regions relevant to the predicted trajectory. To complement rather than duplicate Current evidence, Geometry-first Top2-History selects up to two same-episode cached observations at least 33 synchronized points before the anchor, without appearance similarity or a learned router. Starting from Current coverage, its score combines valid in-frame footprint, age decay, and incremental trajectory-ROI coverage. It accepts the highest positive-gain candidate, updates the coverage union, and repeats once, so the second source must add support; unsupported candidates have zero gain. Selected latent slices form one History canvas with slotwise maximum coverage. Thus, History contains zero, one, or two causal observations. Appendix~\ref{app:c2m} gives the exact score and placement contract.

Geometric support specifies where evidence is admissible, but not how Child-domain features should be expressed in the Mother expert. Pose alignment gives source $q\in\{\mathrm{cur},\mathrm{hist}\}$ a value canvas $V_s^q$ and coverage $C_s^q$ at Mother slot $s$, with mask $m_s^q=\mathbf 1[C_s^q\ge\tau]$. For future Mother group $k\in\{1,2\}$, the translator produces $T_{k,s}^{q,\ell}=\mathcal E_{\mathrm{app}}^\ell(V_s^q)$ and the zero-initialized projection produces $Q_{k,s}^{q,\ell}=\mathcal Z_M^\ell(T_{k,s}^{q,\ell})$. Let $b_s=(m_s^{\mathrm{cur}},m_s^{\mathrm{hist}})$. At a fixed $(k,s,\ell)$, write $R=R_{C\rightarrow M,k,s}^{\ell}$, $Q_{\mathrm{cur}}=Q_{k,s}^{\mathrm{cur},\ell}$, and $Q_{\mathrm{hist}}=Q_{k,s}^{\mathrm{hist},\ell}$, and denote the sigmoid gate predicted from $\bar X_{M,k,s}^{\ell}$ by $\gamma\in[0,1]$. The coverage gate writes
\begin{equation}
R=\begin{cases}
0, & b_s=(0,0),\\
Q_{\mathrm{cur}}, & b_s=(1,0),\\
Q_{\mathrm{hist}}, & b_s=(0,1),\\
\gamma Q_{\mathrm{cur}}+(1-\gamma)Q_{\mathrm{hist}}, & b_s=(1,1).
\end{cases}
\label{eq:c2m_residual}
\end{equation}
Geometry therefore licenses each write: unsupported slots remain strict Null, single-source slots use the available translated residual, and dual-source slots blend residuals only after both sources are licensed. The trust gate cannot create support, leaving the Mother expert to complete remaining unsupported regions.

\subsection{Learning and Causal Inference}

Training jointly supervises future generation and the geometric conditions that govern residual writes. Both streams apply flow matching \cite{lipman2023flowmatching} to non-anchor latent groups, while training combines generation, pose, and M2C residual-trajectory supervision:
\begin{equation}
\mathcal L=
\lambda_M\mathcal L_{\mathrm{FM}}^M+
\lambda_C\mathcal L_{\mathrm{FM}}^C+
\lambda_P\mathcal L_{\mathrm{pose}}+
\alpha_{\mathrm w}\lambda_{\mathrm{aux}}\mathcal L_{\mathrm{aux}}.
\label{eq:objective}
\end{equation}
The pose loss covers position, orientation, validity, and coarse endpoint motion. Stop-gradient protects the geometric interface while flow matching trains how each route uses it; Mother flow matching directly supervises C2M without an appearance proxy. Loss decompositions and weights are given in Appendix~\ref{app:implementation}.

At inference, the schedule enforces the same causal evidence boundary. At each denoising step, routing modules read the pre-injection snapshot, and Pose Readout uses the Mother partition. Current uses its predicted anchor pose and each earlier History observation its cached pose, so inference needs no ground-truth pose. C2M admits only $z_C^0$ and earlier observations as appearance evidence; future tokens remain internal denoising variables. Routes are recomputed as latents evolve, and both anchors are restored after each scheduler update; Appendices~\ref{app:implementation} and~\ref{app:pose} provide the corresponding routing configuration and pose diagnostics.

\section{Experiments}

\subsection{Experimental Setup}

\paragraph{Training details.}
All systems use rank-128 LoRA \cite{hu2022lora} for 10,000 steps on the same split. CrossScope assigns full- and half-width 30-layer Wan2.2-TI2V-5B DiTs to Child and Mother, respectively, initializing the Mother by interpolation; the experts use separate adapters and a shared frozen VAE. Each window subsamples 33 synchronized points at stride four into one anchor and eight targets. Phantom tests use 860 causal windows from 15 episodes; real-world evaluation uses a patient- and case-disjoint 45/7 split within one cohort. We report PSNR, SSIM, FID, and LPIPS \cite{heusel2017gans,zhang2018unreasonable} alongside frozen task evaluators applied identically to generated and target clips. Appendices~\ref{app:evaluation} and~\ref{app:implementation} detail metric aggregation and implementation.

\paragraph{Baselines.}
We compare HunyuanVideo-I2V, Wan2.2, and Cosmos-H-Surgical \cite{kong2024hunyuanvideo,wan2025wan22,he2025cosmossurgical} as independently fine-tuned, view-specific controls, alongside Early-Fusion Wan2.2 and Symmetric Cross-Attention MoT (Symmetric MoT). The first three predict each view independently, whereas Early-Fusion tests direct input coupling. Symmetric MoT retains separate experts but uses mirror-symmetric, zero-initialized cross-attention residuals without Pose Readout, M2C descriptors, or C2M placement. The comparison thus separates independent prediction, generic coupling, and directional routing. All complete systems use the same paired windows, observed anchors, prompt, optimization budget, and evaluation protocol. Table~\ref{tab:main_comparison} reports end-to-end results, and Table~\ref{tab:ablation} gives the same-backbone route analysis. Appendices~\ref{app:implementation} and~\ref{app:evaluation} provide the corresponding configuration and evaluator details.

\begin{figure*}[!t]
    \centering
    \includegraphics[width=0.88\textwidth]{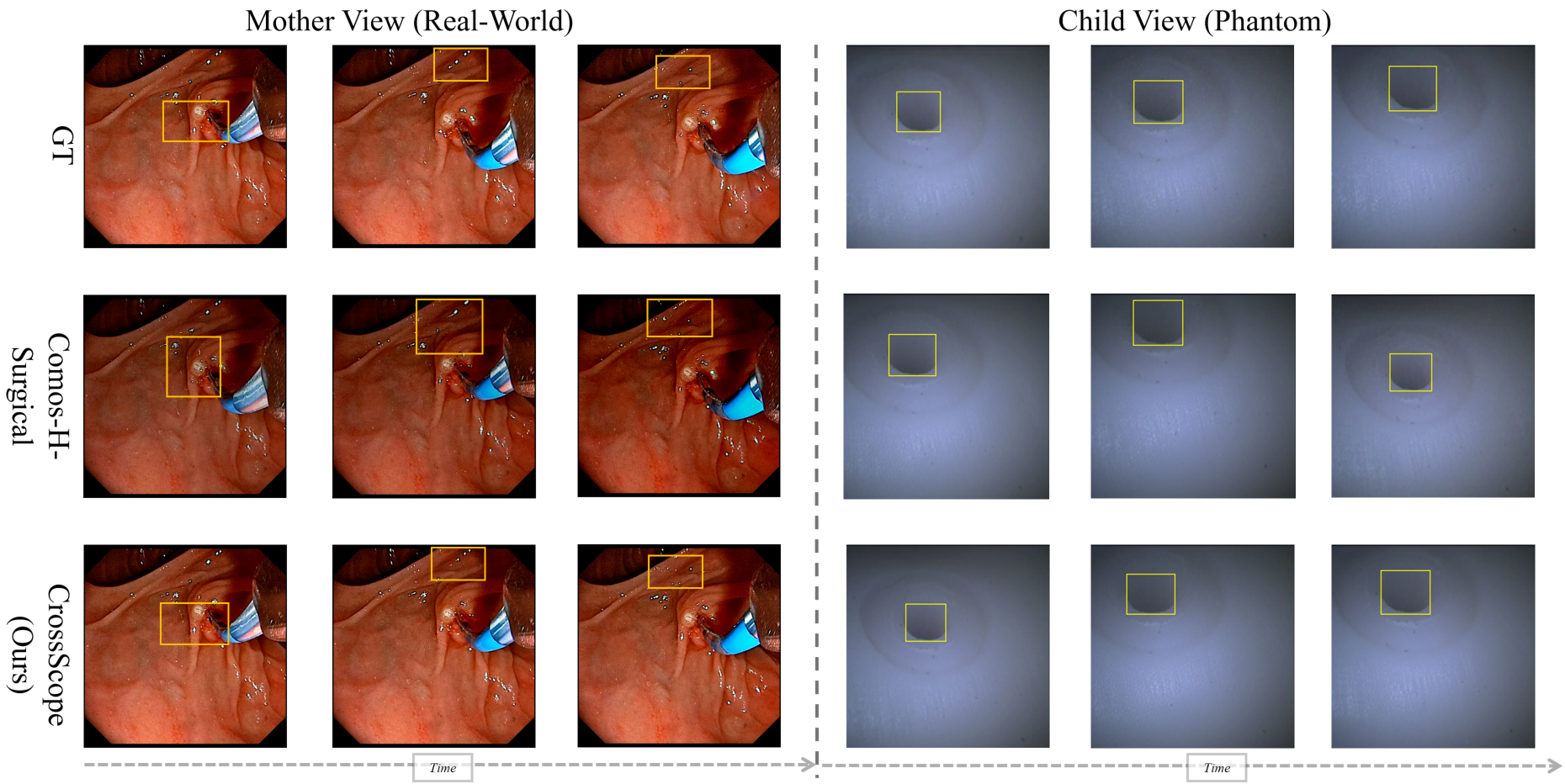}
    \caption{Qualitative future prediction across views. Independent examples are shown for a real-world Mother sequence (left) and a phantom Child sequence (right). Rows show ground truth, Cosmos-H-Surgical, and CrossScope; yellow boxes mark comparison regions.}
    \label{fig:qualitative_results}
\end{figure*}

\subsection{Main Results}

\paragraph{View-specific generation fidelity.}
The main comparison asks whether independent prediction, input fusion, symmetric exchange, or directional routing best preserves two visually distinct targets. Table~\ref{tab:main_comparison}(a) combines distortion measures (PSNR and SSIM) with perceptual measures (FID and LPIPS), requiring agreement with target frames and their appearance statistics. The independent baselines maintain competitive Child appearance, whereas Early-Fusion and Symmetric MoT concentrate their gains more strongly in the Mother view and do not consistently preserve Child fidelity. Thus, increasing cross-view exchange alone does not determine which evidence is useful to each target. CrossScope yields the strongest reported overall profile among the evaluated systems: relative to Cosmos-H-Surgical, it reduces Mother FID by $16.0\%$ and improves Child PSNR and SSIM, while retaining low Child perceptual errors. This role-balanced pattern is consistent with separate view experts and target-specific residual exchange.

\paragraph{Task fidelity and motion.}
Frame fidelity alone does not establish whether dual-scope geometry is preserved. Table~\ref{tab:main_comparison}(b) therefore measures Mother foreground structure (mask IoU), Child localization and visibility (box IoU and recall), and path and terminal motion (centerline and endpoint errors). Symmetric MoT achieves the lowest centerline error among the controls, but its endpoint error remains above Cosmos-H-Surgical, showing that mean path agreement need not recover the final motion state. CrossScope leads all reported task metrics, reducing endpoint error by $33.0\%$ relative to Cosmos-H-Surgical while improving centerline alignment and localization coverage. The gains thus extend from appearance to spatial structure and motion without an observed frame-fidelity trade-off.

\begin{table}[!t]
    \centering
    \begingroup
    \small
    \setlength{\tabcolsep}{3.0pt}

    % ==================== Child View ====================
    \begin{tabular*}{\linewidth}{
        @{\extracolsep{\fill}}lcccc@{}
    }
        \toprule
        \multicolumn{1}{c}{
            \multirow{2}{*}{\textbf{Model}}
        }
        &
        \multicolumn{4}{c}{
            \textbf{Child View}
        }
        \\
        \cmidrule(lr){2-5}
        &
        PSNR$\uparrow$
        & SSIM$\uparrow$
        & FID$\downarrow$
        & LPIPS$\downarrow$
        \\
        \midrule

        MoT-only
        & 34.868
        & 0.939
        & 17.959
        & 0.0855
        \\

        CrossScope-M2C
        & \underline{35.113}
        & \underline{0.941}
        & \underline{17.534}
        & 0.0838
        \\

        CrossScope-C2M
        & 34.927
        & \underline{0.941}
        & 17.633
        & \underline{0.0831}
        \\

        \textbf{CrossScope}
        & \textbf{35.906}
        & \textbf{0.951}
        & \textbf{17.402}
        & \textbf{0.0819}
        \\
        \bottomrule
    \end{tabular*}

    \par

    % ==================== Mother View ====================
    \begin{tabular*}{\linewidth}{
        @{\extracolsep{\fill}}lcccc@{}
    }
        \toprule
        \multicolumn{1}{c}{
            \multirow{2}{*}{\textbf{Model}}
        }
        &
        \multicolumn{4}{c}{
            \textbf{Mother View}
        }
        \\
        \cmidrule(lr){2-5}
        &
        PSNR$\uparrow$
        & SSIM$\uparrow$
        & FID$\downarrow$
        & LPIPS$\downarrow$
        \\
        \midrule

        MoT-only
        & 35.871
        & 0.965
        & 37.575
        & 0.0356
        \\

        CrossScope-M2C
        & 35.934
        & 0.967
        & 37.200
        & 0.0356
        \\

        CrossScope-C2M
        & \underline{36.542}
        & \underline{0.969}
        & \underline{31.277}
        & \underline{0.0343}
        \\

        \textbf{CrossScope}
        & \textbf{37.147}
        & \textbf{0.973}
        & \textbf{28.232}
        & \textbf{0.0334}
        \\
        \bottomrule
    \end{tabular*}

    \par
    \setlength{\tabcolsep}{0.8pt}

    % ==================== Task Fidelity and Motion ====================
    \begin{tabular*}{\linewidth}{
        @{\extracolsep{\fill}}lccccc@{}
    }
        \toprule

        % ================= Group headers =================
        \multicolumn{1}{c}{
            \multirow{2}{*}{\textbf{Model}}
        }
        &
        \multicolumn{1}{c}{
            \shortstack{
                \textbf{Mother}\\
                \textbf{Structure}
            }
        }
        &
        \multicolumn{2}{c}{
            \shortstack{
                \textbf{Child}\\
                \textbf{Localization}
            }
        }
        &
        \multicolumn{2}{c}{
            \shortstack{
                \textbf{Child}\\
                \textbf{Trajectory}
            }
        }
        \\

        \cmidrule(lr){2-2}
        \cmidrule(lr){3-4}
        \cmidrule(lr){5-6}

        % ================= Metric headers =================
        &
        \shortstack{Mask\\IoU$\uparrow$}
        &
        \shortstack{Box\\IoU$\uparrow$}
        &
        \shortstack{Det.\\Recall$\uparrow$}
        &
        \shortstack{End.\\Err.$\downarrow$}
        &
        \shortstack{Cent.\\Err.$\downarrow$}
        \\
        \midrule

        MoT-only
        & 0.943
        & 0.793
        & 0.919
        & 26.593
        & 13.727
        \\

        CrossScope-M2C
        & 0.943
        & \underline{0.816}
        & \underline{0.923}
        & \underline{21.665}
        & \underline{12.437}
        \\

        CrossScope-C2M
        & \underline{0.947}
        & 0.795
        & 0.919
        & 25.537
        & 13.426
        \\

        \textbf{CrossScope}
        & \textbf{0.948}
        & \textbf{0.839}
        & \textbf{0.927}
        & \textbf{19.343}
        & \textbf{11.238}
        \\
        \bottomrule
    \end{tabular*}

    \endgroup

    \caption{Route-level directional ablation on the phantom benchmark.
    The three blocks report Child frame fidelity, Mother frame fidelity,
    and task fidelity/motion; trajectory errors are in pixels. Bold and
    underlined values mark the best and second-best results.}
    \label{tab:ablation}
\end{table}

\subsection{Ablation Study}

\paragraph{Directional specialization.}
Table~\ref{tab:ablation} asks whether the gains follow the intended direction of each route within the same MoT backbone and optimization schedule. Adding M2C concentrates its largest improvements on Child localization and trajectory quality, reducing endpoint error by $18.5\%$ over MoT-only while leaving Mother fidelity nearly unchanged. C2M shows the complementary profile: it improves every Mother frame-fidelity measure and reduces Mother FID by $16.8\%$, while its Child-motion changes remain smaller. This selective response matches the architecture: pose-derived transition descriptors write motion residuals only to the Child expert, whereas pose-aligned appearance evidence writes only to the Mother expert. The alignment supports route-specific functional specialization rather than a uniform effect of generic cross-view communication across the two experts.

\begin{figure}[!ht]
    \centering
    \includegraphics[width=\linewidth]{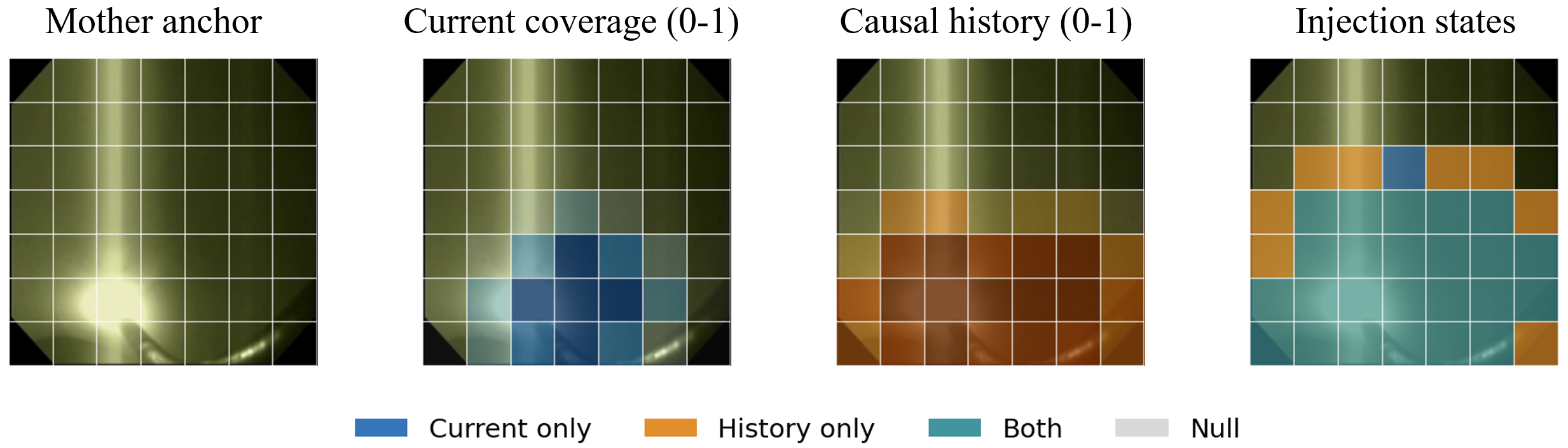}
    \caption{C2M spatial-support diagnostic.
    A held-out phantom window from the 10,000-step C2M-only diagnostic. Columns show the Mother anchor, Current and causal-History coverage, and the resulting $7\times7$ injection states; additional cases appear in Appendix Figure~\ref{fig:app_c2m_evidence}.}
    \label{fig:c2m_support_main}
\end{figure}

\paragraph{C2M spatial-support behavior.}
Figure~\ref{fig:c2m_support_main} isolates how C2M forms admissible Mother-plane writes. Current coverage remains localized around the anchor-aligned region, whereas selected causal History contributes complementary support. Their composition yields Current-only, History-only, and dual-source states while retaining Null elsewhere, matching the geometry-licensed residual contract in Equation~\ref{eq:c2m_residual}.

\paragraph{Route complementarity.}
If the routes were redundant, the full model would match the stronger single-route variant for each target. Instead, adding C2M to CrossScope-M2C lowers Mother FID from $37.200$ to $28.232$, while adding M2C to CrossScope-C2M lowers Child endpoint error from $25.537$ to $19.343$ pixels. C2M supplies Mother appearance evidence absent from motion-only routing, whereas M2C corrects Child displacement left by appearance-only routing. Their combination improves both role-specific metric families rather than reallocating fidelity between streams.

\subsection{Real-World Evaluation}

\begin{figure}[ht]
    \centering
    \begingroup
    \setlength{\tabcolsep}{2pt}
    \begin{tabular}{@{}cc@{}}
        \multicolumn{2}{c}{\includegraphics[width=0.6\columnwidth]{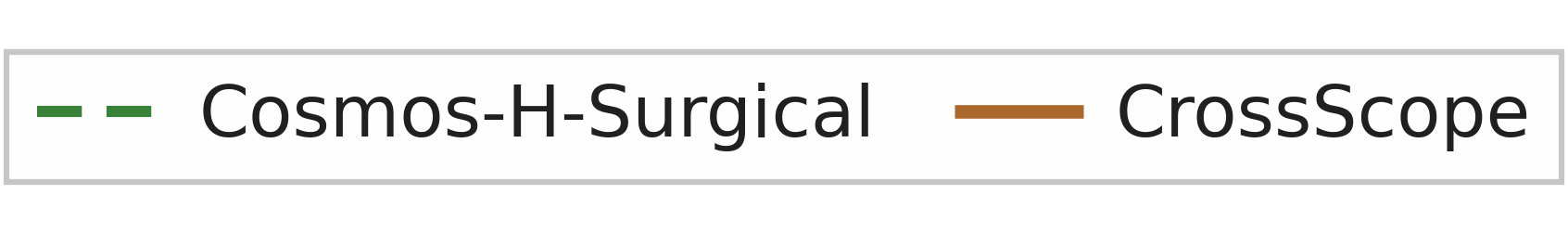}} \\[2pt]
        \includegraphics[width=0.46\columnwidth]{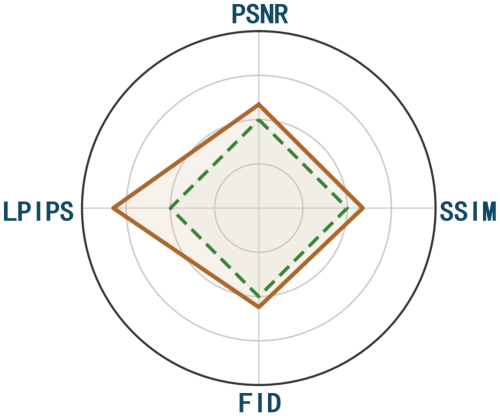} &
        \includegraphics[width=0.46\columnwidth]{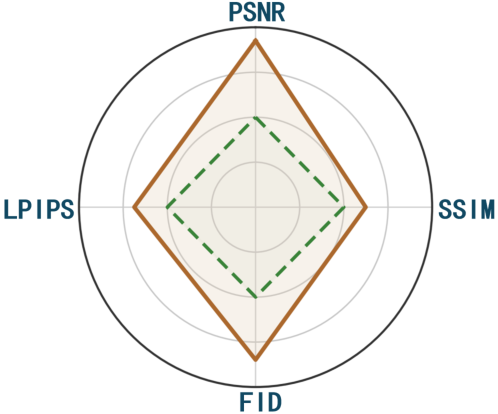} \\
        \textbf{(a)} Child view & \textbf{(b)} Mother view \\[4pt]
        \includegraphics[width=0.46\columnwidth]{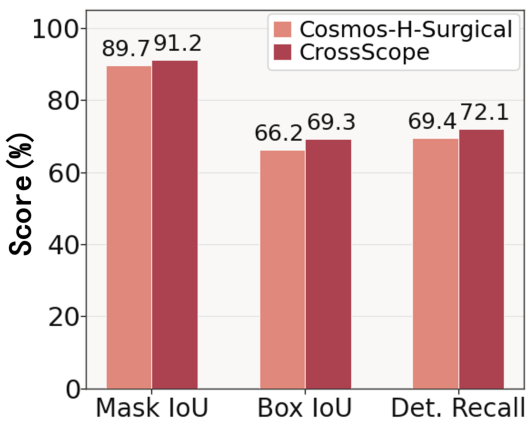} &
        \includegraphics[width=0.46\columnwidth]{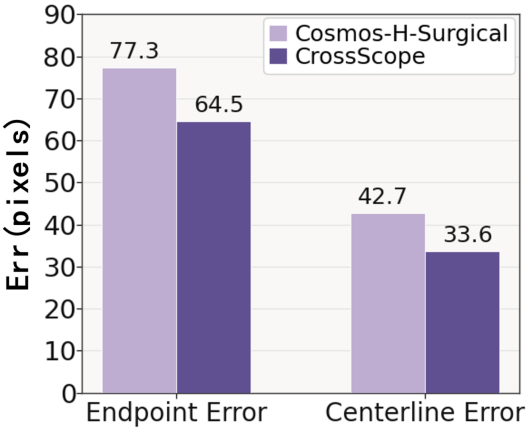} \\
        \textbf{(c)} Task fidelity & \textbf{(d)} Motion quality
    \end{tabular}
    \endgroup
    \caption{Real-world quantitative comparison on seven held-out episodes. Radar scores are normalized to Cosmos-H-Surgical ($=100$), using reciprocal ratios for lower-is-better metrics. Bars report task-fidelity scores (\%) and trajectory errors (pixels). Exact values are provided in Appendix~\ref{app:evaluation}.}
    \label{fig:real_world_quantitative}
\end{figure}

\paragraph{Held-out quantitative performance.}
We test whether phantom trends persist under real-world appearance, illumination, and scope-motion variation. We fine-tune Cosmos-H-Surgical and CrossScope on 45 real episodes and evaluate seven patient- and case-disjoint episodes from the same cohort. Across the role-specific measures in Figure~\ref{fig:real_world_quantitative}, Mother PSNR increases by $1.57\,\mathrm{dB}$, while endpoint and centerline errors decrease by $16.5\%$ and $21.5\%$. Consistent gains across both roles match the phantom pattern. Appendix~\ref{app:evaluation} reports exact values and metric definitions.

\paragraph{Qualitative comparison.}
In Figure~\ref{fig:qualitative_results}, CrossScope more closely preserves broad-context local structure in the Mother example and follows near-field lumen position and appearance in the Child example than Cosmos-H-Surgical. These independent sequences visually contextualize the structure, localization, and motion trends reported by the aggregate quantitative evaluation.

\paragraph{Cross-setting evidence.}
Together, complete-system comparisons distinguish directional routing from independent prediction, input fusion, and symmetric exchange; same-backbone ablations link each route to its target; and real-world evaluation tests the joint pattern under natural acquisition variation. Concurrent gains in Mother fidelity and structure and in Child fidelity, localization, and motion indicate a role-consistent forecast rather than an isolated metric change. View-specific measures test whether routing preserves both predictive roles rather than reallocating fidelity between streams.

\section{Conclusion}

CrossScope structures dual-scope forecasting around what to share, which target receives it, and where to write it. In Mother--Child ERCP, role-preserving experts use a Mother-plane trajectory for Child motion and admit pose-aligned Child appearance to Mother prediction only under geometric support. Across phantom and held-out real-world episodes in the paired benchmark, CrossScope improves the reported measures of frame fidelity, structure, localization, and motion; same-backbone ablations associate each route with its intended role. These results support target-specific evidence routing as a design principle for paired flexible scopes.

The present study is limited to short-horizon prediction in one dual-scope procedure, with real-world evaluation drawn from one cohort. Future work will extend CrossScope to longer horizons, additional dual-scope procedures, and prospective validation across centers and devices before considering clinical decision-support applications.

\begingroup
\small
\bibliography{aaai2027}
\endgroup

\onecolumn
\setcounter{secnumdepth}{2}
\setcounter{section}{0}
\renewcommand{\thesection}{\Alph{section}}

\section{Additional Method Details}
\label{app:method}

This supplement documents the methodological and empirical contracts underlying CrossScope. Section~A specifies the information each route may read and the residuals it may write. Section~B records the training and inference configuration. Section~C details the paired benchmark and its view-specific annotations. Section~D defines the evaluation protocol and reports exact real-world values. Section~E examines the intermediate pose interface and output-level Child motion. Together, these details connect the directional design in Figure~\ref{fig:architecture} to the measurements reported in the main paper.

\subsection{Read--Write Routing Invariant}

At every routed transformer layer, CrossScope separates evidence reads from residual writes. Let $H_M$ and $H_C$ denote the pre-injection Mother and Child states. The manuscript term \emph{Joint Tokens} denotes the conceptual paired snapshot $[H_M;H_C]$ available for reading; it is not a third mutable stream. Pose Readout, M2C, and C2M therefore inspect the same pre-write state, but add their outputs only to their designated expert. Neither route can condition on the other route's same-layer residual. This invariant gives the paths independent write semantics: M2C converts a pose-derived transition description into a Child residual, whereas C2M converts pose-aligned observed Child evidence into a Mother residual. Zero-initialized output projections make both paths exact no-ops at initialization, preserving the uncoupled view experts before the routes are learned.

\subsection{Mother-to-Child Motion Conditioning}

\paragraph{Pose Readout interface.}
Pose Readout is attached at layer 7. It uses nine learned temporal queries with sinusoidal temporal embeddings and a pooled flow-matching timestep embedding. The queries cross-attend to memory formed from the pre-injection Mother tokens and pinned Mother anchor-token group. Its inference inputs are therefore the Mother state, observed Mother anchor, and denoising timestep; it neither reads Child tokens nor consumes ground-truth pose. The output is a nine-state Mother-plane sequence
$(x,y,\cos\theta,\sin\theta,\ell_v)$. Position is bounded to $[-0.5,0.5]^2$, heading is normalized to the unit circle, and $\ell_v$ is a validity logit. Before M2C or C2M uses the prediction, the route-facing pose is detached and invalid states are masked by $\ell_v\leq0$. This prevents generation losses from redefining the geometric interface while leaving the pose objective to supervise it directly.

\paragraph{Sampling-aware descriptor and residual write.}
M2C transforms the detached pose sequence into a descriptor for each valid transition between temporally sampled states. The per-frame pose cache stores $(x,y,\cos\theta,\sin\theta,q)$, where $q$ is validity. Derivatives are formed after temporal sampling, rather than on the raw synchronized stream, so velocity and acceleration use the interval represented by the generated video. Table~\ref{tab:app_motion_descriptor} lists the twelve components. The four ego-motion components form the compact coarse condition, whereas the complete descriptor supplies fine transition-wise evidence.

\begin{table}[ht]
    \centering
    \appTableStyle
    \setlength{\tabcolsep}{4pt}
    \begin{tabular}{C{0.25\linewidth}L{0.65\linewidth}}
        \toprule
        \multicolumn{1}{c}{Group} & \multicolumn{1}{c}{Components and interpretation} \\
        \midrule
        Global pose & $(x_k,y_k,\cos\theta_k,\sin\theta_k)$: normalized tip position and unit heading at the transition start. \\
        Ego motion & $(\Delta x_k^{\mathrm{ego}},\Delta y_k^{\mathrm{ego}},\sin\Delta\theta_k,\cos\Delta\theta_k)$: translation expressed in the current tip frame and wrap-safe heading change. \\
        First order & $(v_k,\omega_k)$: translational speed and angular velocity over the sampled interval. \\
        Second order & $(a_k,\alpha_k)$: finite differences of $v_k$ and $\omega_k$; both are zero for the first transition. \\
        \bottomrule
    \end{tabular}
    \normalsize
    \caption{Twelve-dimensional Mother-plane transition descriptor used by M2C.}
    \label{tab:app_motion_descriptor}
\end{table}

For consecutive sampled poses, let $\Delta x_t=x_{t+1}-x_t$ and $\Delta y_t=y_{t+1}-y_t$. Translation is rotated into the current tip frame before it is encoded:
\[
\begin{bmatrix}
\Delta x_t^{\mathrm{ego}}\\
\Delta y_t^{\mathrm{ego}}
\end{bmatrix}
=
\begin{bmatrix}
\cos\theta_t & \sin\theta_t\\
-\sin\theta_t & \cos\theta_t
\end{bmatrix}
\begin{bmatrix}
\Delta x_t\\
\Delta y_t
\end{bmatrix}.
\]
The heading change is represented by $(\sin\Delta\theta_t,\cos\Delta\theta_t)$ rather than direct angle subtraction, avoiding wrap discontinuities at $\pm\pi$. Translational speed and angular velocity are computed over the sampled interval, and their finite differences produce $a_t$ and $\alpha_t$. The first transition uses zero second-order terms. Hence, the nine predicted states yield eight fine transition descriptors. The two coarse conditions are recomputed directly from endpoint pairs $(0,4)$ and $(4,8)$, rather than by averaging fine transitions; they condition the first and second future latent groups, respectively.

At its enabled Child layers, M2C first writes a validity-masked cross-attention residual from the fine descriptors, then an AdaLN residual conditioned on the corresponding coarse descriptor. The anchor latent group is unchanged. A missing pose state is not replaced by a nominal motion vector; its transition contributes no M2C residual. Fine conditions retain transition-level variation, whereas the two coarse conditions provide the group-level context used by AdaLN. The route remains disabled during the first 3,000 optimization steps, allowing the view experts and Pose Readout to establish stable single-view representations before motion residuals are introduced. Its auxiliary decoder reads the residual fields written at these layers rather than the raw Pose Readout output, so the auxiliary signal supervises the generation pathway it conditions.

\subsection{C2M Causal Appearance Placement}
\label{app:c2m}

C2M has a different information contract from M2C: it writes only pose-aligned appearance from observed Child frames to the Mother expert. The Current source is the observed Child anchor; History consists solely of strictly earlier cached Child observations. Future Child-token states remain internal denoising variables and are never admitted as C2M appearance evidence. The nine predicted poses are grouped into anchor state $0$, future states $1$--$4$, and future states $5$--$8$, matching the three VAE latent groups. Current uses the observed anchor latent and its group-$0$ pose. For each historical observation, its latent slices are placed with their cached poses before evidence is aggregated across slices. This ordering prevents a pose--appearance mismatch from aggregating temporal evidence before spatial alignment.

\begin{figure}[ht]
    \centering
    \includegraphics[width=0.82\linewidth]{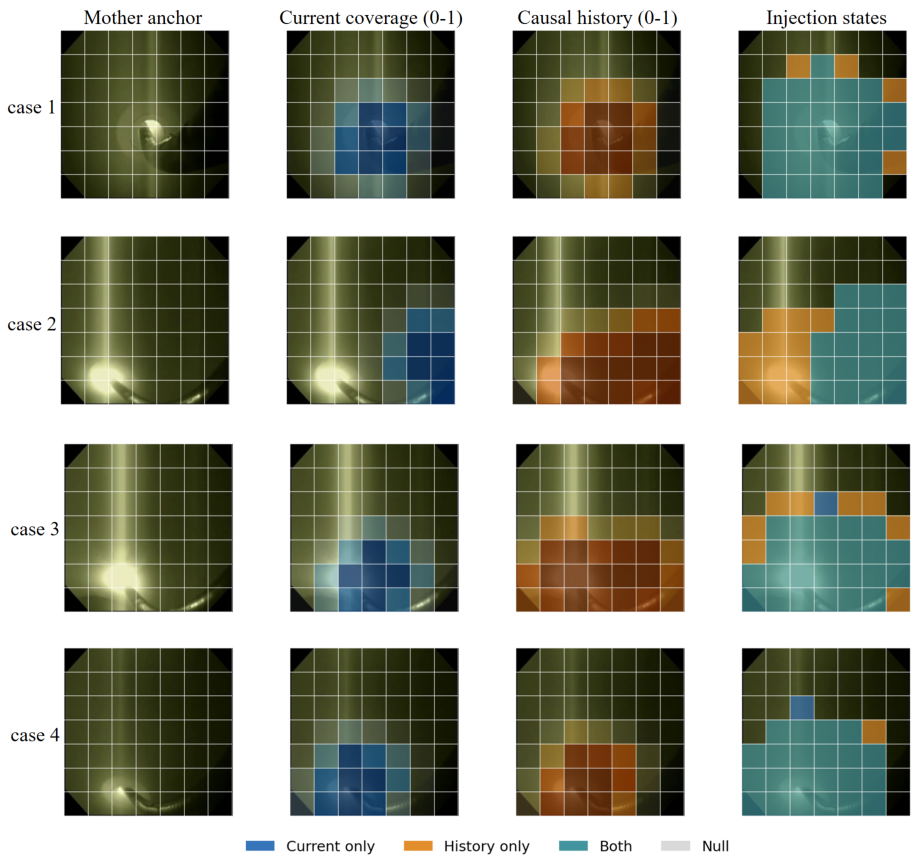}
    \caption{Geometry-licensed C2M evidence placement.
    Four held-out phantom windows from the 10,000-step C2M-only geometry-canvas ablation are visualized at the final denoising step. Columns show the Mother anchor, continuous Current coverage, continuous causal-History coverage, and the final discrete $7\times7$ injection state. Blue, orange, teal, and gray denote Current-only, History-only, dual-source, and Null support, respectively. This is a route diagnostic, not a complete-system performance result.}
    \label{fig:app_c2m_evidence}
\end{figure}

Figure~\ref{fig:c2m_support_main} highlights one representative window in the main paper, while Figure~\ref{fig:app_c2m_evidence} extends this diagnostic to four held-out phantom windows. Together, they separate continuous geometric coverage, which licenses a source at a slot, from the discrete write state, which determines whether the translated residual is written from Current, History, both, or neither.

Each admissible source is projected to a $7\times7$ Mother-plane canvas. The projection produces an in-frame footprint and a coverage value for every Mother slot. C2M then selects complementary History evidence by geometry, not by an appearance-similarity score. For a historical candidate $j$, let $C_s^j$ be its coverage at slot $s$, $f_j$ its in-frame footprint ratio, $a_j$ its age in synchronized frames, and $C$ the coverage union already obtained from Current and previously selected History. Its marginal gain is

\begin{equation}
G_j(C)=f_j\,2^{-a_j/h}
\sum_{k=1}^{2}\sum_s\rho_{k,s}
\left[C_s^j-C_s\right]_+,
\qquad [x]_+=\max(x,0),
\label{eq:history_gain}
\end{equation}
where $h$ is the recency half-life and $\rho_{k,s}$ weights slot $s$ by the predicted-trajectory region of interest for future group $k$. Selection starts from Current coverage. After the highest positive-gain candidate is selected, its support is merged into the union before the second round. The procedure thus returns zero, one, or two causal History observations, and the second observation must add support not already supplied by Current or the first History observation.

The score acts only on geometric coverage and recency; it does not inspect an appearance-similarity feature. Selection therefore determines \emph{where} History can contribute before the learned translator determines \emph{how} its values should be expressed in the Mother expert. If no candidate has positive marginal gain, the History canvas remains empty and C2M reduces to Current-only or Null behavior according to available support.

Geometry establishes support before any C2M residual is written. At every enabled Mother layer, Current and History values are translated \emph{independently} into Mother-domain residuals. A Mother-state-conditioned sigmoid gate blends these translations only at slots supported by both sources. At single-source slots, the available translation is written directly; at unsupported slots, the update is exactly zero. The learned gate therefore chooses how to combine already licensed sources, but cannot create support where geometry provides none. The canvas size, layer schedule, and warm-up are recorded in Table~\ref{tab:app_configuration}.

\section{Implementation and Reproducibility}
\label{app:implementation}

\subsection{Training and Routing Configuration}

All reported systems use the fixed prompt ``ERCP dual-scope cannulation phase,'' the same paired training split, prediction horizon, and evaluation windows. Table~\ref{tab:app_configuration} records the configuration for the reported 10,000-step CrossScope runs. It distinguishes the common optimization protocol from settings specific to the two directional routes, allowing the route architecture to be reproduced without inferring implementation values from the main text.

Training was performed on one Ubuntu 4XN6lS-80GB node with four NVIDIA H100 GPUs (80~GB each), 88 vCPU cores, and 960~GiB of system memory, using Python~3.10.16, PyTorch~2.7.1, and CUDA~12.8. Each reported result is obtained from one training run per model using training seed 42 and fixed generation seed 12,345.

\begin{table}[ht]
    \centering
    \appTableStyle
    \setlength{\tabcolsep}{4pt}
    \begin{tabular}{C{0.31\linewidth}L{0.59\linewidth}}
        \toprule
        \multicolumn{1}{c}{Setting} & \multicolumn{1}{c}{Value} \\
        \midrule
        Backbone and VAE & Wan2.2-TI2V-5B; 30-layer Child/Mother DiTs with hidden/FFN dimensions $3072/14336$ and $1536/7168$; interpolated Mother initialization, independent rank-128 LoRA adapters, and a shared frozen WanVideoVAE38. \\
        Input and adaptation & $224\times224$ per view; bfloat16; per-device batch size 16; no gradient accumulation. \\
        Optimizer and schedule & AdamW with $(\beta_1,\beta_2)=(0.9,0.95)$, learning rate $10^{-4}$, zero weight decay, and cosine decay. \\
        Optimization and sampling & 10,000 optimization steps; training/split seed 42; fixed generation seed 12,345 and 20 denoising steps at evaluation. \\
        Pose and M2C & Pose Readout at layer 7; M2C layers $[8,10,12,14,16,18,20,22,25,28]$; 3,000-step M2C warm-up. \\
        C2M & Layers $[8,11,14,17,20,23,26,29]$; $7\times7$ canvas; 500-step Null warm-up; footprint scale $0.1$; kernel width $0.15$. \\
        Causal history & Coverage threshold $\tau=0.05$; recency half-life $h=200$ frames; history candidates precede the Current anchor by at least 33 frames. \\
        \bottomrule
    \end{tabular}
    \normalsize
    \caption{Training, routing, and causal-history settings used for the reported CrossScope runs.}
    \label{tab:app_configuration}
\end{table}

\subsection{Baselines, Capacity, and Objectives}

Every reported model is LoRA-fine-tuned for 10,000 optimization steps. HunyuanVideo-I2V, Wan2.2, and Cosmos-H-Surgical use one view-specific instance for each target stream. Early-Fusion Wan2.2 encodes paired anchors on a horizontally concatenated $224\times448$ canvas and predicts the paired future in that layout. Symmetric Cross-Attention MoT retains separate Mother and Child latents and experts, but replaces directional routing with mirror-symmetric, zero-initialized cross-attention residuals. It has neither Pose Readout nor the M2C descriptor and C2M geometry contracts. MoT-only disables all three CrossScope additions, whereas CrossScope-M2C and CrossScope-C2M activate only the named path. Before the common evaluator is applied, every output is separated into its two views.

CrossScope contains $688.3$ million trainable parameters, including the LoRA adapters, Pose Readout, M2C, and C2M, and excluding frozen VAE and base-backbone weights. This reports the trainable capacity of the complete CrossScope implementation. The directional attribution in the main paper comes from the MoT-only and single-route variants, which retain the same dual-stream backbone and optimization schedule.

The main objective combines Mother and Child flow matching, pose supervision, and an M2C residual-trajectory auxiliary term. Its outer weights are
\[
\lambda_M=1,\qquad \lambda_C=1,\qquad
\lambda_P=1,\qquad \lambda_{\mathrm{aux}}=0.1.
\]
The Pose Readout objective is
\[
\mathcal L_{\mathrm{pose}}=
\mathcal L_{xy}+0.5\mathcal L_{\mathrm{dir}}+
0.05\mathcal L_{\mathrm{unit}}+0.2\mathcal L_{\mathrm{valid}}+
\mathcal L_{\mathrm{coarse}},
\]
which supervises normalized position, heading, direction-vector norm, validity, and coarse endpoint motion. The M2C auxiliary objective is
\[
\mathcal L_{\mathrm{aux}}=
\mathcal L_{\mathrm{traj}}+\mathcal L_{\mathrm{endpoint}}+
0.5\mathcal L_{\mathrm{ROI}}+0.2\mathcal L_{\mathrm{valid}}.
\]
The auxiliary decoder reads the M2C residual fields after they have been constructed, rather than the raw pose output. Mother flow matching directly supervises C2M's appearance translator and its support-conditioned residual writes.

\subsection{Causal Inference Schedule}

Table~\ref{tab:app_causal_schedule} records the information available at each inference-stage operation. Predicted pose is a geometric condition. Future Child latent states may participate in the Child denoising stream, but they are never admitted as C2M appearance evidence; C2M receives only the observed anchor and strictly earlier cached observations.

\begin{table}[ht]
    \centering
    \appTableStyle
    \setlength{\tabcolsep}{4pt}
    \begin{tabular}{C{0.21\linewidth}L{0.30\linewidth}L{0.42\linewidth}}
        \toprule
        \multicolumn{1}{c}{Stage} & \multicolumn{1}{c}{Available information} & \multicolumn{1}{c}{Operation} \\
        \midrule
        Anchor preparation & Observed Mother and Child anchors; cached strictly earlier Child observations & Pin the two observed anchors and retain only causal Child history for potential C2M evidence. \\
        Pose Readout & Pre-injection Mother state and Mother anchor tokens & Predict the nine Mother-plane pose states and detach the route-facing pose representation. \\
        M2C update & Detached pose transitions and Child pre-injection state & Write fine motion and coarse AdaLN residuals only to future Child token groups. \\
        C2M update & Observed Child anchor, causal Child history, predicted poses, and Mother pre-injection state & Establish support, translate Current and History evidence, and write only geometry-licensed Mother residuals. \\
        Scheduler step & Updated future latents and the pinned anchors & Recompute routes as latents evolve, then restore both observed anchors. \\
        \bottomrule
    \end{tabular}
    \normalsize
    \caption{Causal information flow during CrossScope inference.}
    \label{tab:app_causal_schedule}
\end{table}

At rollout start, the observed Mother and Child anchors are VAE-encoded independently and copied into latent group $0$, while the two future groups are initialized from noise. Both experts use the same scheduler timestep. After each scheduler update, the saved anchor latents are restored before the next denoising step. Frame $0$ is therefore never denoised as a prediction; it remains the observed condition while future latents and route conditions are recomputed throughout inference.

\section{Dual-Scope Benchmark and Annotation Protocol}
\label{app:benchmark}

\subsection{Benchmark Composition and Split}

Each synchronized time point is a paired Mother--Child observation: the wide-view Mother and near-field Child frames arise at the same procedural instant while the two scopes remain independently controlled. The phantom collection contains 300 paired episodes (285 training and 15 test), and the real-world collection contains 52 (45 training and seven held-out test). Accordingly, the time-point counts in Table~\ref{tab:app_benchmark_composition} refer to synchronized pairs, not individual view frames. Splits are episode-disjoint; real-world test episodes are patient- and case-disjoint. Experienced ERCP clinicians reviewed pairing, synchronization, and task-specific annotations.

At the model boundary, the two views remain separate tensors rather than a horizontally concatenated canvas. Each raw $480\times360$ stream is independently resized to $224\times224$. A 33-point synchronized clip is then sampled at an interval of four to form the nine-frame input--target sequence. This preserves the view-specific image statistics seen by the two experts while retaining a shared temporal reference for paired prediction and evaluation.

\begin{table}[ht]
    \centering
    \appTableStyle
    \setlength{\tabcolsep}{4pt}
    \begin{tabular}{C{0.17\linewidth}C{0.17\linewidth}C{0.19\linewidth}L{0.35\linewidth}}
        \toprule
        \multicolumn{1}{c}{Source} & \multicolumn{1}{c}{Episodes} & \multicolumn{1}{c}{Synchronized time points} & \multicolumn{1}{c}{Paired annotations} \\
        \midrule
        Phantom & 285 train / 15 test & 596,190 & Mother segmentation map and mask-derived Child-tip pose; Child papilla box. \\
        Real-world & 45 train / 7 test & 34,800 & Mother segmentation map and mask-derived Child-tip pose; Child papilla box. \\
        \bottomrule
    \end{tabular}
    \normalsize
    \caption{Composition and view-specific annotations of the Dual-Scope Benchmark. Each synchronized time point contains one paired Mother--Child observation.}
    \label{tab:app_benchmark_composition}
\end{table}

\subsection{View-Specific Annotation Pipeline}

\begin{figure}[ht]
    \centering
    \includegraphics[width=\linewidth]{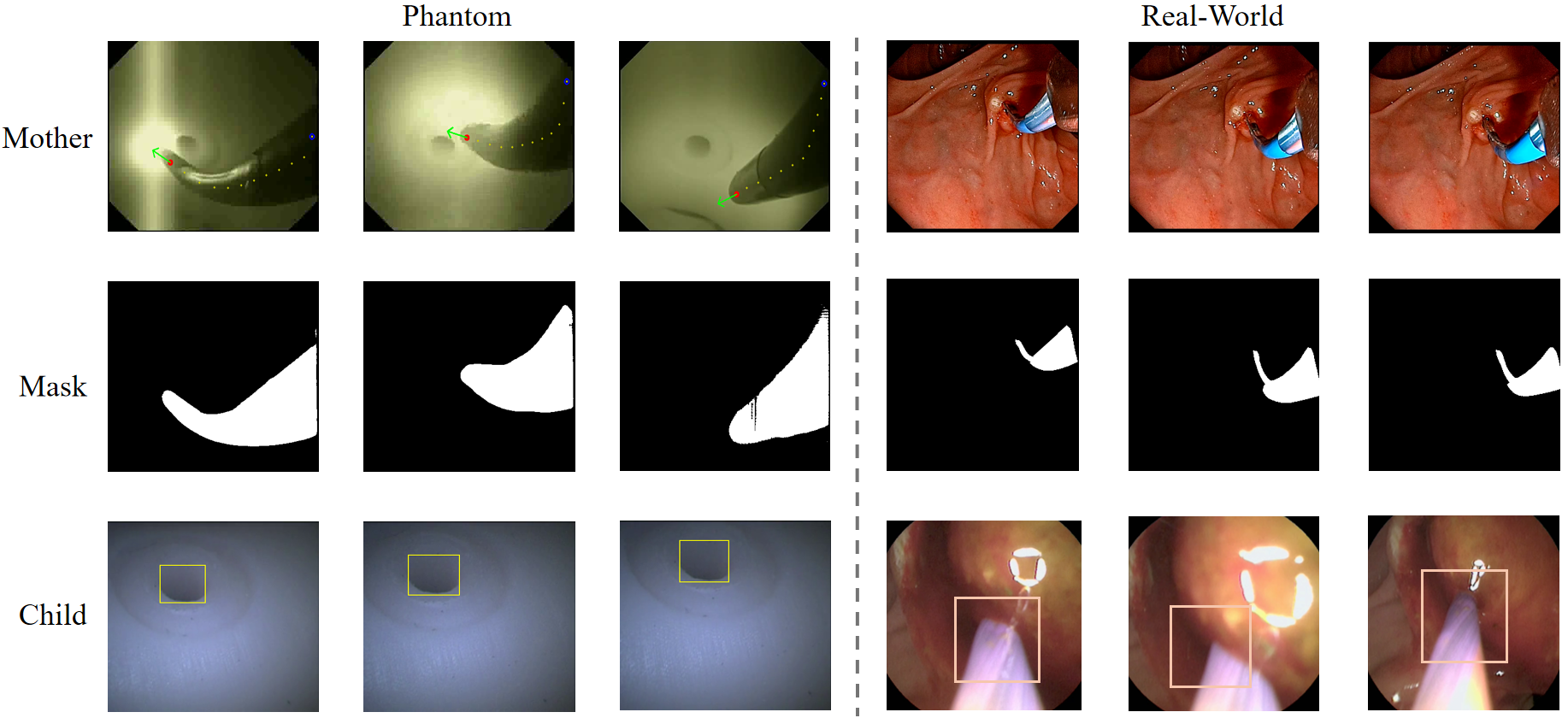}
    \caption{View-specific annotations in the Dual-Scope Benchmark. Each column is a synchronized Mother--mask--Child triplet; phantom and real-world examples are shown. In the Mother row, yellow points trace the mask-derived centerline, the blue circle marks the base, the red point marks the tip, and the green arrow marks its heading. The middle row shows the Mother segmentation map, and the Child row shows the papilla bounding box.}
    \label{fig:app_annotations}
\end{figure}

Figure~\ref{fig:app_annotations} links paired raw observations to the task-specific quantities used throughout the paper. In the Mother view, the segmentation map separates background from the Child-endoscope foreground. A mask-derived extractor records a normalized tip center, unit heading vector, and validity label for each frame. The pose cache is created before temporal sampling, allowing the same per-frame geometric primitive to be differentiated at the interval of a training or evaluation window. These mask-derived poses supervise Pose Readout and provide the Mother-plane geometric reference for both directional routes.

The pose extractor follows a deterministic image-plane procedure. It retains the largest valid foreground component, estimates its centerline by binning points along the principal axis, and assigns the endpoint nearest an image boundary as the instrument base. The opposite endpoint is the tip, and the local tangent near that tip is oriented from base to tip. A frame that fails the geometric validity check is marked invalid rather than assigned a synthetic pose. This makes the visual overlays in Figure~\ref{fig:app_annotations} directly traceable to the Mother segmentation map.

In the Child view, a papilla bounding box specifies target visibility and localization. The fixed detector is applied independently to generated and target clips under the same protocol. Its box centers provide the anchored displacements used for centerline and endpoint errors, whereas box overlap and recall measure localization and coverage. Thus, the paired annotation chain supports Mother structure, Child target visibility, and Child trajectory fidelity in both phantom and real-world episodes.

\section{Evaluation Protocol and Exact Results}
\label{app:evaluation}

\subsection{Prediction Windows and Frame Metrics}

Each prediction instance begins with 33 consecutive synchronized time points, temporally sampled into nine frames per view. Frame $0$ is the observed anchor, and frames $1$--$8$ are future prediction targets. The phantom test split yields 860 non-overlapping windows from 15 held-out episodes; real-world metrics are computed on seven held-out episodes. Window construction, prediction horizon, denoising budget, and generation seed are fixed across methods.

Frame-fidelity metrics assess only generated future frames. PSNR, SSIM, and AlexNet LPIPS are averaged over the eight predicted frames of each view. FID uses standard InceptionV3 pool3 features, pooling generated and target future frames separately for Mother and Child. By contrast, task-specific evaluation retains the anchor when it provides the reference required to measure a trajectory. Table~\ref{tab:app_metric_sources} maps each reported metric family to its view, temporal support, and fixed evaluator.

\begin{table}[ht]
    \centering
    \appTableStyle
    \setlength{\tabcolsep}{4pt}
    \begin{tabular}{C{0.24\linewidth}C{0.14\linewidth}C{0.18\linewidth}L{0.34\linewidth}}
        \toprule
        \multicolumn{1}{c}{Metric family} & \multicolumn{1}{c}{View} & \multicolumn{1}{c}{Temporal support} & \multicolumn{1}{c}{Evaluation source} \\
        \midrule
        PSNR, SSIM, FID, LPIPS & Mother, Child & Eight future frames & Generated and target frames. \\
        Mask IoU & Mother & Anchor plus eight future frames & Fixed Child-endoscope foreground segmenter. \\
        Box IoU and recall & Child & Anchor plus eight future frames & Fixed papilla detector applied to generated and target clips. \\
        Centerline and endpoint error & Child & Common detected frames in the nine-frame sequence & Anchored generated and target box-center displacements. \\
        \bottomrule
    \end{tabular}
    \normalsize
    \caption{Relationship between reported metric families, temporal support, and fixed evaluation sources.}
    \label{tab:app_metric_sources}
\end{table}

\subsection{Structure, Localization, and Motion Aggregation}
\label{app:task_metrics}

Mother structure is evaluated with a binary DeepLabV3-MobileNetV3-Large student trained on phantom Mother frames with SAM2-propagated Child-endoscope masks. Child localization is evaluated with a YOLO11x detector trained on a labeled ERCP Child corpus for papillary-orifice, lumen, and calculus detection. Its highest-confidence papillary-orifice candidate, at a confidence threshold of $0.25$, supplies the reported localization and trajectory measurements. Both checkpoints are frozen. Generated and target clips receive identical preprocessing: each square model frame is restored to the native $480\times360$ aspect ratio before detection. Task-specific quantities use the full conditioned nine-frame sequence. The anchor is included in mask and box measurements for consistent conditioned-sequence evaluation, and establishes the displacement reference for trajectory metrics. Target-side detections define the reference objects that generated clips are expected to reproduce.

The aggregates distinguish coverage from geometric accuracy. Detection recall is defined over all target-side papilla boxes, so a missing generated detection is a false negative. Box IoU is computed only for matched generated and target boxes, and trajectory errors require at least two common detections. Thus, an absent box is not converted into an arbitrary pixel penalty, while recall remains visible alongside conditional localization and trajectory scores. Table~\ref{tab:app_evaluator_denominators} states these denominators explicitly.

\begin{table}[ht]
    \centering
    \appTableStyle
    \setlength{\tabcolsep}{3.5pt}
    \begin{tabular}{C{0.25\linewidth}L{0.65\linewidth}}
        \toprule
        \multicolumn{1}{c}{Metric} & \multicolumn{1}{c}{Aggregation and missing-detection semantics} \\
        \midrule
        Mask IoU & All nine Mother frames; generated and target Child-endoscope foreground masks are scored framewise. \\
        Detection recall & All target-side papilla boxes; a miss is a false negative. \\
        Box IoU & Matched target-side/generated papilla boxes; a miss has no IoU value. \\
        Trajectory errors & Papilla sequences with at least two common detections; no fixed error is inserted for a miss. \\
        \bottomrule
    \end{tabular}
    \normalsize
    \caption{Evaluator denominators and missing-detection semantics.}
    \label{tab:app_evaluator_denominators}
\end{table}

For the papilla target detected in both clips, let $c_t$ and $c_t^\star$ be the generated and target box centers at common frames, and let $t_0$ be their first common frame. We compare anchored displacements $d_t=c_t-c_{t_0}$ and $d_t^\star=c_t^\star-c_{t_0}^\star$. Centerline error is the mean $\lVert d_t-d_t^\star\rVert_2$ over common frames, whereas endpoint error is the same quantity at the final common frame. The first measures path alignment across the available horizon; the second exposes accumulated terminal drift. Reporting recall alongside these conditional errors keeps coverage and matched-trajectory accuracy interpretable as complementary quantities.

\subsection{Exact Real-World Results}

The main paper summarizes the two-model real-world comparison with baseline-relative radar axes and bar plots. Table~\ref{tab:app_real_exact} provides the corresponding unnormalized values, so that the frame-fidelity, Mother-structure, Child-localization, and Child-motion values underlying Figure~\ref{fig:real_world_quantitative} can be inspected directly.

\begin{table}[ht]
    \centering
    \appTableStyle
    \setlength{\tabcolsep}{2.8pt}
    \begin{tabular*}{\linewidth}{@{\extracolsep{\fill}}cccccc@{}}
        \toprule
        Model & View & PSNR$\uparrow$ & SSIM$\uparrow$ & FID$\downarrow$ & LPIPS$\downarrow$ \\
        \midrule
        Cosmos-H-Surgical & Child & 18.314 & 0.737 & 19.153 & 0.445 \\
        Cosmos-H-Surgical & Mother & 18.453 & 0.726 & 37.240 & 0.360 \\
        CrossScope & Child & \textbf{18.627} & \textbf{0.750} & \textbf{18.925} & \textbf{0.418} \\
        CrossScope & Mother & \textbf{20.023} & \textbf{0.744} & \textbf{34.813} & \textbf{0.347} \\
        \bottomrule
    \end{tabular*}
    \smallskip
    \begin{tabular*}{\linewidth}{@{\extracolsep{\fill}}cccccc@{}}
        \toprule
        Model & Mask$\uparrow$ & Box$\uparrow$ & Recall$\uparrow$ & End.$\downarrow$ & Cent.$\downarrow$ \\
        \midrule
        Cosmos-H-Surgical & 0.897 & 0.662 & 0.694 & 77.283 & 42.732 \\
        CrossScope & \textbf{0.912} & \textbf{0.693} & \textbf{0.721} & \textbf{64.535} & \textbf{33.557} \\
        \bottomrule
    \end{tabular*}
    \normalsize
    \caption{Exact real-world results on seven held-out episodes. In the lower panel, Mask is Mother Child-endoscope mask IoU; Box and Recall are Child papilla-box IoU and recall; End. and Cent. are endpoint and centerline errors in pixels.}
    \label{tab:app_real_exact}
\end{table}

The table preserves the two view roles rather than averaging them into one score. Its upper panel reports frame fidelity on each stream's native scale, and its lower panel retains the structural, localization, coverage, and motion quantities used by the task evaluators. The main-paper radar is a compact relative visualization; Table~\ref{tab:app_real_exact} supplies absolute values and metric directions without cross-metric normalization. All entries use the same seven held-out episodes, fixed generation protocol, and evaluator definitions described above.

\section{Pose Readout and Generated Motion Diagnostics}
\label{app:pose}

\subsection{Pose Readout Diagnostics}

Pose Readout is evaluated at optimization step 10,000 on 860 non-overlapping held-out phantom windows. The nine predicted pose states in each window are compared with Mother-plane targets derived from segmentation maps. Position errors are converted from normalized coordinates to the resized image plane, and heading error is the angular distance between predicted and target direction vectors. Table~\ref{tab:app_pose_diagnostics} reports position, heading, coarse-motion, and full-descriptor errors.

The diagnostics address two route-facing interfaces. Absolute position and heading characterize the Mother-plane geometry used for C2M placement. Coarse four-dimensional and raw twelve-dimensional errors characterize the derived transition quantities consumed by M2C after temporal sampling. Reporting both avoids treating a low absolute-pose error as evidence of accurate motion differences.

\begin{table}[ht]
    \centering
    \appTableStyle
    \setlength{\tabcolsep}{5pt}
    \begin{tabular}{cc}
        \toprule
        Diagnostic & Value \\
        \midrule
        Normalized position L1 & 0.02795 \\
        Per-coordinate position MAE & $6.250$ px \\
        Euclidean position error & $9.750$ px \\
        Heading error & $5.406^\circ$ \\
        Coarse-motion 4D L1 & 0.02194 \\
        Raw-motion 12D L1 & 0.02450 \\
        \bottomrule
    \end{tabular}
    \normalsize
    \caption{Pose Readout diagnostics on 860 held-out phantom windows.}
    \label{tab:app_pose_diagnostics}
\end{table}

\begin{figure}[ht]
    \centering
    \includegraphics[width=\linewidth]{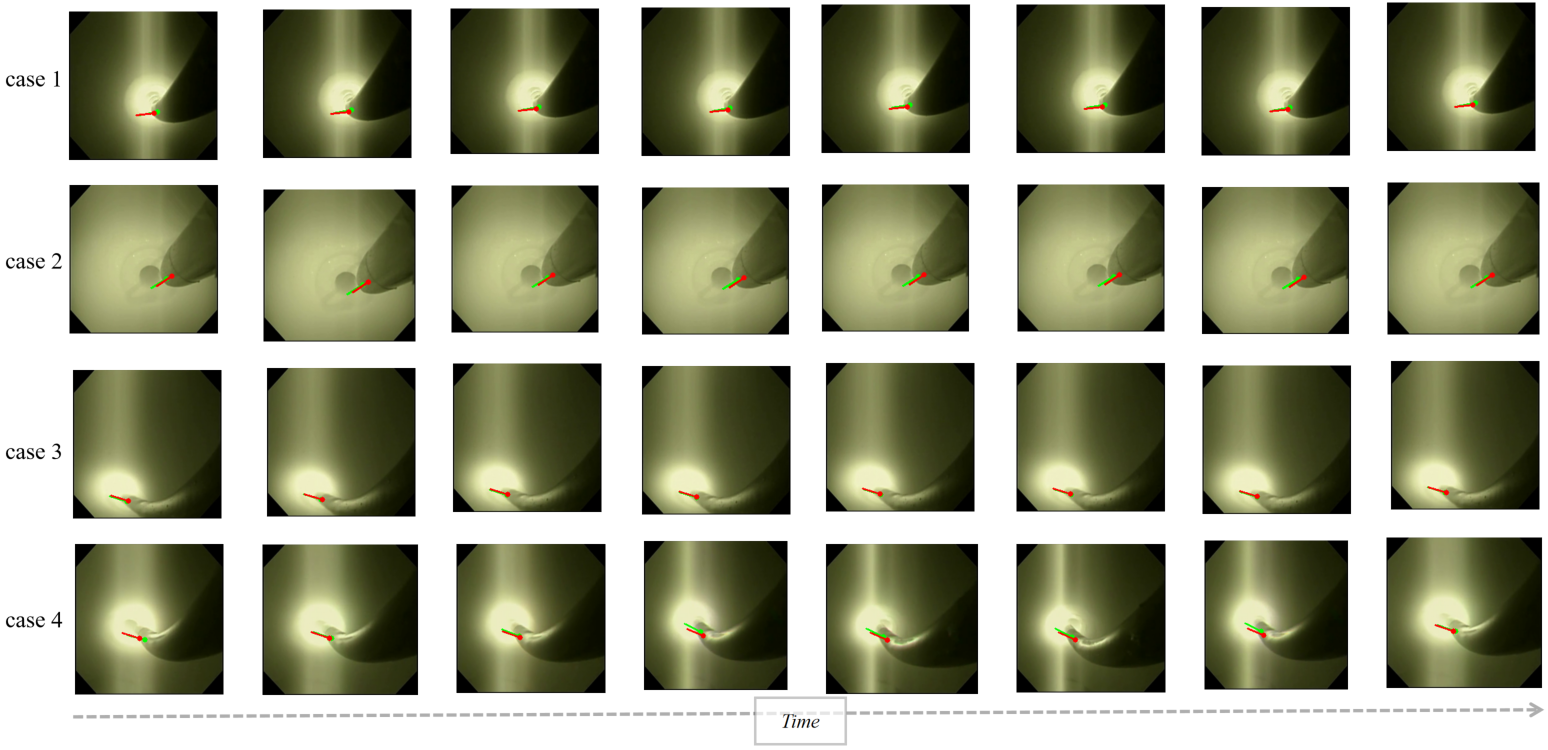}
    \caption{Visual validation of Pose Readout.
    Four held-out phantom windows at step 10,000 are shown over their nine temporally sampled Mother states. Red markers and rays denote Pose Readout predictions; green markers and rays denote mask-derived targets.}
    \label{fig:app_pose_overlays}
\end{figure}

Figure~\ref{fig:app_pose_overlays} complements the aggregate diagnostics with state-level overlays of predicted and mask-derived pose primitives on held-out Mother frames.

The shuffled-target and constant-prior controls test whether the readout exploits a generic location prior rather than the Mother-stream state. Shuffling preserves the marginal target distribution but breaks the correspondence between each window and its pose target; the constant-prior predictor retains no window-specific geometry. Normalized position L1 rises from $0.02795$ to $0.15137$ when targets are shuffled across windows, and is $0.11328$ for a constant-prior predictor. This gap is consistent with recovery of window-specific Mother-plane information. All 7,740 target pose states in this diagnostic are valid, so validity is not separately tabulated.

\subsection{Generated Child Bounding-Box Trajectories}

\begin{figure}[!b]
    \centering
    \includegraphics[width=1\linewidth]{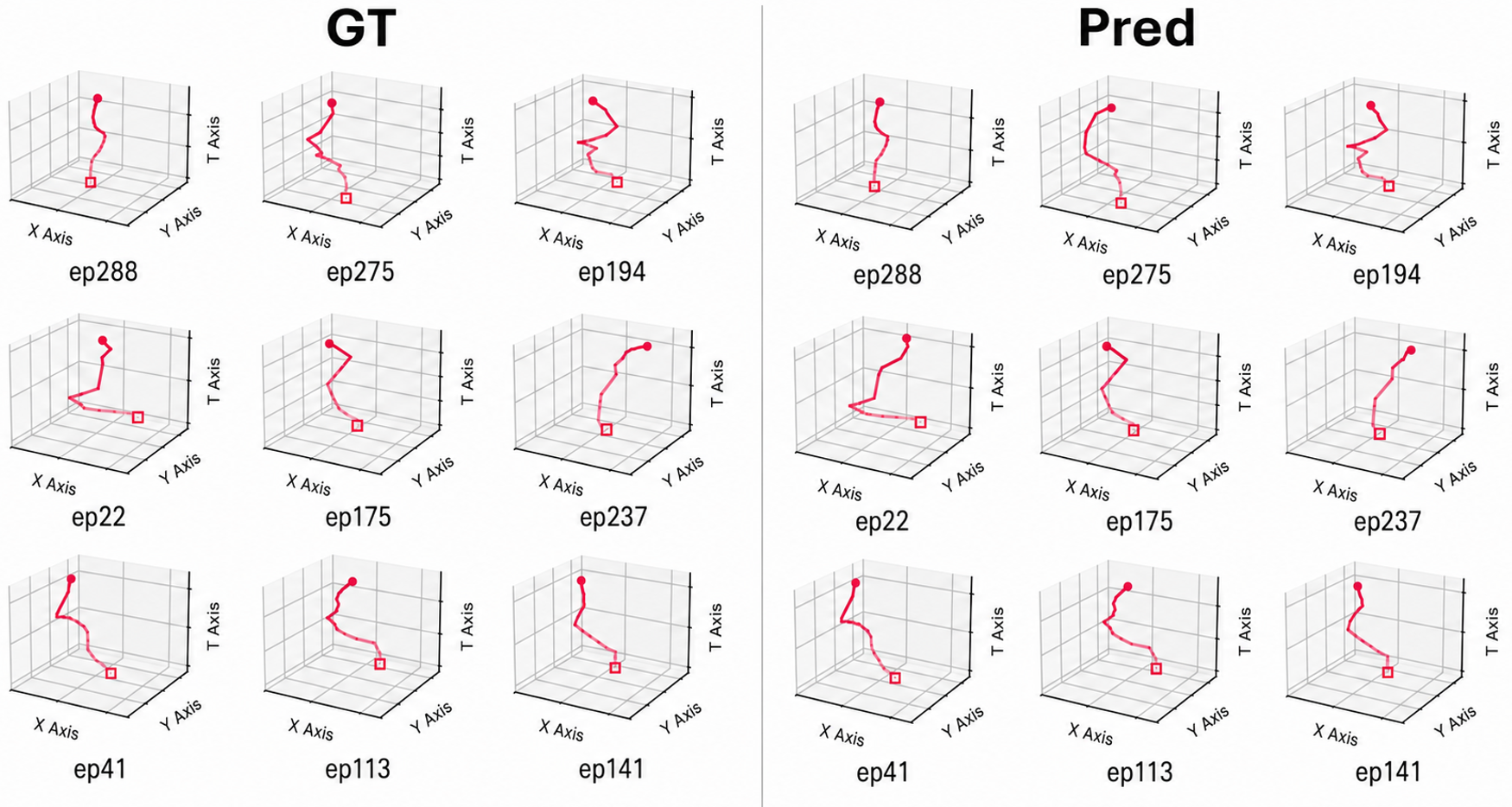}
    \caption{Output-level Child papilla trajectories for nine selected held-out phantom episodes at step 10,000. The left and right blocks compare target and CrossScope-generated mean bounding-box centers from 20 independent nine-frame windows sampled across each episode, in $(X,Y,T)$ space. Open squares and filled circles mark the first and final detections.}
    \label{fig:app_bbox_trajectories}
\end{figure}

The preceding diagnostic assesses the intermediate Mother-plane pose interface. Figure~\ref{fig:app_bbox_trajectories} instead visualizes output-level papilla motion from bounding-box centers independently detected in target and generated Child clips; it does not reuse the Pose Readout target. The $X$ and $Y$ axes are detector-coordinate centers and $T$ is the original frame index, exposing lateral drift or jitter that a two-dimensional overlay can conceal.

For each displayed episode, 20 nine-frame windows are sampled evenly across the full timeline. Each window is generated independently from its observed anchor. Detector centers within a window are averaged into one point placed at the raw center-frame index. A centered five-point moving average is then drawn through the sequence to suppress window-level detector jitter. The left and right panels therefore use matched temporal locations in the target and CrossScope-generated clips, but do not depict an autoregressive chained rollout.

Target and generated clips are passed through the detector independently. A window contributes a point only when the papilla is detected in that clip; no location is interpolated or assigned a default coordinate. Both plots retain the same detector-coordinate system, with no post-hoc spatial alignment. The displayed smoothing shows the episode-level trend only. The main-paper Box, Recall, Centerline, and Endpoint metrics use unsmoothed framewise detections under the rule in Section~\ref{app:task_metrics}. These traces therefore complement the aggregate errors temporally; they are not a test of long autoregressive rollout stability.


\end{document}